\documentclass[11pt]{article}
\usepackage[final]{acl}
\usepackage{times}
\usepackage{latexsym}
\usepackage[T1]{fontenc}
\usepackage[utf8]{inputenc}
\usepackage{hyperref}
\usepackage{xcolor}
\usepackage{booktabs}       
\usepackage[utf8]{inputenc} 
\usepackage[T1]{fontenc}    
\usepackage{url}            
\usepackage{nicefrac}       
\usepackage{microtype}      
\usepackage{makecell}
\usepackage{enumitem}
\usepackage{pifont}
\usepackage{url}
\usepackage{graphicx}
\usepackage{amsmath}
\usepackage{amsfonts}
\usepackage{amssymb}
\usepackage{bbm}
\usepackage{siunitx}
\usepackage{changepage}
\usepackage{multirow}
\usepackage{wrapfig, lipsum}
\usepackage{siunitx}
\DeclareMathOperator*{\argmax}{arg\,max}

\title{Do Language Models Have Beliefs? \\ Methods for Detecting, Updating, and Visualizing Model Beliefs}

\author{
 Peter Hase$^{1,2}$ \ \ \ \ \ \ \ \ \ \
 Mona Diab$^{1}$ \ \ \ \ \ \ \ \ \ \ \
 Asli Celikyilmaz$^{1}$ \ \ \ \ \ \ \ \ \
 Xian Li$^{1}$ \\
 \textbf{Zornitsa Kozareva}$^{1}$ \ \ \ \ \
 \textbf{Veselin Stoyanov}$^{1}$ \ \ \ \ \ \
 \textbf{Mohit Bansal}$^{2}$ \ \ \ \ \ \ \ \ \
 \textbf{Srinivasan Iyer}$^{1}$ \\
 $^{1}$Meta AI~~~~~~
 $^{2}$UNC Chapel Hill \\
 \small\texttt{ \{peter, mbansal\}@cs.unc.edu}\\
 \small\texttt{ \{mdiab, aslic, xianl, zori, ves, sviyer\}@fb.com}
}
 
\begin{document}

\maketitle

\begin{abstract}
Do language models have beliefs about the world? \citet{dennett1995animals} famously argues that even thermostats have beliefs, on the view that a belief is simply an informational state decoupled from any motivational state. 
In this paper, we discuss approaches to detecting when models have beliefs about the world, 
and we improve on methods for updating model beliefs to be more truthful, with a focus on methods based on learned optimizers or hypernetworks. Our main contributions include:
(1) new metrics for evaluating belief-updating methods that focus on the logical consistency of beliefs, (2)
a training objective for Sequential, Local, and Generalizing model updates (SLAG) that improves the performance of learned optimizers,
and (3) the introduction of the \emph{belief graph}, which is a new form of interface with language models that shows the interdependencies between model beliefs. 
Our experiments suggest that models possess belief-like qualities to only a limited extent, but update methods can both fix incorrect model beliefs and greatly improve their consistency. Although off-the-shelf optimizers are surprisingly strong belief-updating baselines, our learned optimizers can outperform them in more difficult settings than have been considered in past work.\footnote{All supporting code for experiments in this paper is publicly available at \url{https://github.com/peterbhase/SLAG-Belief-Updating}.}
\end{abstract}

\section{Introduction}

Language models (LMs) may not have beliefs in the same sense that people do, but there are a few reasons to analyze LMs in terms of the beliefs they may possess. The first is that this is a useful way to understand and speak about how LMs behave. When discussing whether animals have beliefs (raccoons, in particular), philosopher Daniel Dennett \citeyearpar{dennett1995animals} writes:
\begin{adjustwidth}{3mm}{3mm}
\emph{You might as well call the state of the raccoon a belief, since if you call it a ``registration'' or a ``data-structure'' in the ``environmental information store'' or some other technical term, the logic you use to draw inferences about the animal's behavior, given its internal states, will be the standard, ``intentionalistic'' logic of belief.}
\end{adjustwidth}
Dennett bases this conclusion in the fact that we can and do draw accurate inferences about animal behavior by first understanding their beliefs. We are drawn to speak about the beliefs of LMs in the same ``maximally bland (but maximally useful!)'' sense. To the extent that these neural networks act intelligently in response to stimuli, we may form more accurate theories of how they work by understanding their beliefs.

The second reason for ascribing beliefs to language models is that many of the stricter definitions of belief incidentally exclude many real beliefs held by real people. Following \citet{dennett1995animals}, \citet{newen2020ascribe} define a belief as \emph{an informational state decoupled from any motivational state}, and they outline a few additional properties of beliefs, namely that they should (1) be recombinable with motivational states and other informational states and (2) have some minimal structural organization. Further, an entity with beliefs should (1) be sensitive to new information, (2) categorize new beliefs as they develop, and (3) display some kind of logical consistency. These are all properties that come in degrees, and setting the bar too high will exclude many of the statements that people earnestly express to others in their everyday lives. Meanwhile, animals and neural networks alike store information in accordance with these properties to at least some extent. 

We also note that we use the term \emph{belief} rather than \emph{knowledge} as in related work \cite{zhu2020modifying, de2021editing} because we want to analyze beliefs \emph{of} language models rather than knowledge \emph{in} them. LMs may contain a great deal of knowledge \emph{to us}, but in a traditional view of knowledge as Justified True Belief, it is relatively more difficult to say that an LM knows something rather than that it believes it \cite{sep-belief}.

\begin{figure*}[t]
    \centering
    \includegraphics[width=0.87\textwidth]{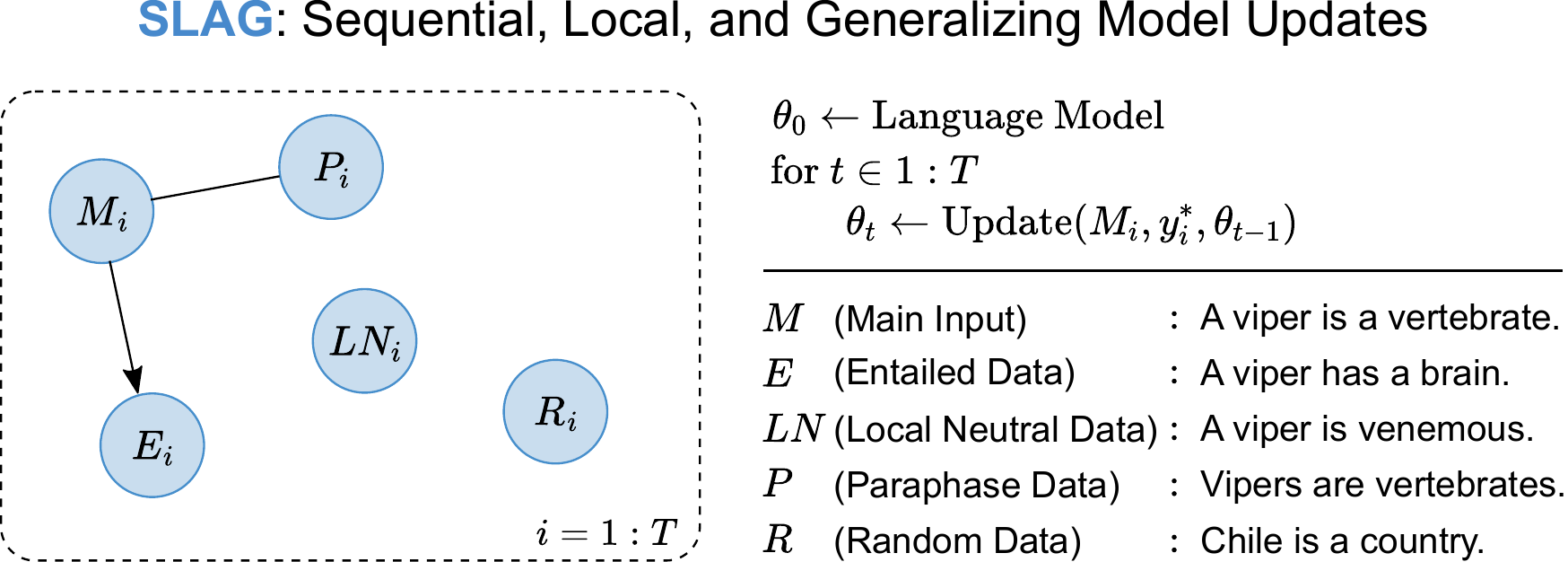}
    \caption{Relying only on a single Main Input $M_i$, we want to make a targeted update to a language model that (1) changes the output for  input $M_i$ to a desired label $y_i^*$ (e.g. True/False, or an answer to a question), (2) changes the output for equivalent paraphrases of $M_i$, (3) appropriately changes outputs for data entailed by the tuple ($M_i, y_i^*$), and (4) does \emph{not} change outputs for other logically neutral data, even if it is similar (local) in some way.}
    \label{fig:main-fig}
    \vspace{-7pt}
\end{figure*}

In the remainder of this paper, we turn our attention to three practical endeavors: \emph{detecting}, \emph{updating}, and \emph{visualizing} beliefs in LMs. 
We build on work on editing models after training, which
is an exciting recent direction of research with many potentially valuable use cases \cite{sinitsin2020editable, zhu2020modifying, de2021editing, mitchell2021fast}. 
For LMs, uses include correcting factually inaccurate outputs and preventing otherwise unwanted outputs from models (e.g. toxic generated text) without expensive data curation and retraining efforts. These are important applications given that LMs (1) struggle with future data when trained on data from the past \cite{lazaridou2021pitfalls, dhingra2021time}, (2) generate morally undesirable text in many situations \cite{gehman2020realtoxicityprompts, bender2021dangers}, and (3) simply give inaccurate outputs for tasks like question answering \cite{lin2021truthfulqa}. Notably, there is good evidence that scaling models to larger sizes will not fix these particular problems or may even exacerbate them in cases like imitative falsehoods in QA, so we will likely need an alternative solution \cite{lazaridou2021pitfalls, gehman2020realtoxicityprompts, lin2021truthfulqa}. 
We next outline a few key contributions of the paper. Figure \ref{fig:main-fig} represents the core ideas behind these contributions. 

\vspace{2pt}
\noindent\textbf{Detecting beliefs.} We measure the degree to which LMs exhibit several properties of belief-possessing systems, using models finetuned on fact verification and question answering tasks.
Beyond simply checking individual model responses, we want to assess the structural properties of model outputs: Are they consistent under paraphrase? Are they logically consistent? Does changing one belief correctly change other entailed beliefs? Does it erroneously change other unrelated beliefs? 
Past work has focused primarily on consistency under paraphrase \cite{elazar2021measuring, de2021editing, mitchell2021fast}. Here, we adapt data from \citet{talmor2020leap} to measure consistency under entailment (including for contrapositives), and we use the Wikidata5m dataset \cite{wang2021kepler} to construct logically neutral belief pairs for checking that models do treat the beliefs as independent.

\vspace{2pt}
\noindent\textbf{Updating beliefs.} We propose a Sequential, Local, and Generalizing belief update objective (SLAG) that substantially improves the performance of the \textsc{KnowledgeEditor} method from \citet{de2021editing}.
\textsc{KnowledgeEditor} is a learned optimizer that edits a model's weights in order to change its prediction on an input while satisfying other desiderata, like consistency under paraphrase. 
Principally, we use more difficult training data for the learned optimizer, and we also train the network to apply multiple small edits rather than just one edit.
These changes markedly improve the overall update success rate and lower the rate at which other beliefs are corrupted. 
Moreover, we find that \textsc{KnowledgeEditor} almost totally fails when updating multiple beliefs in a row as opposed to a changing a single belief.
In this setting, off-the-shelf optimizers are far preferable methods.
However, by explicitly training the optimizer to update multiple beliefs sequentially, we are able to once again outperform off-the-shelf optimizers. 
Lastly, we advocate that these methods be evaluated for their ability to correct false or morally undesirable model beliefs, rather than to arbitrarily adjust model beliefs to plausible alternatives as in past work \cite{zhu2020modifying, de2021editing, mitchell2021fast}.

\vspace{2pt}
\noindent\textbf{Visualizing belief graphs.} We explore a new form of interface with LMs, the \emph{belief graph}. 
Given a set of beliefs, we construct belief graphs by changing each model belief and checking what other beliefs are sensitive to those changes. Each belief becomes a node, and directed edges between nodes show that updating one belief changes the other.
We discuss graph metrics that help summarize the dependencies between model beliefs.

We summarize our main conclusions as follows:
\begin{enumerate}[nosep, wide=0pt, leftmargin=*, after=\strut]
    \item  ${\small \sim}$100M parameter models exhibit limited belief-like qualities, as paraphrase consistency scores are under 70\%, and models show mixed levels of consistency under entailment (Sec. \ref{sec:experiment_detection}). 
    \item Off-the-shelf optimizers are surprisingly effective baselines for updating model beliefs, and they generally outperform learned optimizers when updating a single belief (Sec. \ref{sec:experiment_updating}).
    \item When updating multiple beliefs in a row, 
    method performance greatly declines (especially for learned optimizers).
    By using SLAG, we can improve learned optimizers' performance beyond what baselines can reach (Sec. \ref{sec:experiment_updating}).
    \item Belief graphs reveal many nonsensical dependencies between model beliefs. We find that (1) updates are mostly likely to change already incorrect model beliefs and (2) there are highly connected beliefs which influence a large fraction of all model beliefs (Sec. \ref{sec:belief_graphs}).
\end{enumerate}

\section{Related Work}

\paragraph{Detecting beliefs in language models.}

Much past work has explored how information is stored and represented in pretrained language models \cite{rogers-etal-2020-primer}, though few discuss what qualifies information as a model belief. \citet{petroni-etal-2019-language} provide evidence that LMs store relational information between entities, and \citet{roberts-etal-2020-much} show that LMs can answer open-ended questions. Subsequent work has explored how much knowledge is stored in LMs \cite{heinzerling-inui-2021-language}, approaches to querying models for knowledge \cite{hewitt2019designing, jiang2020can, voita2020information, west2021symbolic}, and methods for learning more knowledge during pretraining \cite{wang2021kepler, wang2021k}. Most relevant to our work are studies from \citet{talmor2020leap} and \citet{elazar2021measuring}. 
\citet{talmor2020leap} train LMs to perform True/False classification of factual claims, and they measure how a model's belief in one fact correlates with its belief in an entailed fact. We use their LeapOfThought dataset to measure model consistency under entailment before and after updating the up-stream beliefs in models. Meanwhile, \citet{elazar2021measuring} measure the consistency of model predictions for sets of paraphrased inputs. We adopt their metric for paraphrase consistency as a measure of belief. 
In concurrent work, \citet{kassner2021beliefbank} discuss consistency under entailment and paraphrase as conditions for belief, and they measure consistency under entailment with a new dataset, BeliefBank.

\paragraph{Updating beliefs in language models.}

Approaches to making targeted updates to model beliefs vary along a few dimensions. First is whether the methods alter model training or operate in a post-training setting. 
\citet{sinitsin2020editable} use a meta-learning objective during training to encourage ease of editing afterwards, though the memory requirements of their approach limit its scalability beyond 100M parameter models. 
A larger family of methods make post-training updates to models, differing in how they formalize the update problem: \citet{dai2021knowledge} propose a hand-crafted algorithm for updating model weights, while \citet{zhu2020modifying} use projected gradient descent for batches of points. \citet{de2021editing} and \citet{mitchell2021fast} frame the problem as a machine learning problem and train hypernetworks (learned optimizers) that process model gradients in order to produce a new model that (1) gives the desired output for the edited point, while (2) incorporating other objectives like minimizing the changes in predictions for other data. Here, we build directly upon the method from \citet{de2021editing}, showing where it fails and providing an improved training objective (SLAG). In particular, we find that the method struggles with updating multiple beliefs sequentially. This setting bears some commonality to the continual learning problem, though continual learning methods generally aim to learn new tasks or datasets rather than make targeted updates to specific model beliefs \cite{parisi2019continual}.

Not all methods edit model weights. \citet{kassner2021beliefbank} update model beliefs by adding in relevant information to the input at test time (to improve consistency under entailment). But as with retrieval-based methods, this approach does not change the model weights and hence does not influence model outputs on all other potentially relevant inputs \cite{lewis2020retrieval, hase2021can}.

\begin{table*}[t]
\begin{center}
\small
\begin{tabular}{l l l p{0.26\textwidth}}
\toprule 
Dataset & Data Type & Input & Label(s)  \\
\midrule
\multirow{2}{*}{zsRE}
& Main Input & Player Ali Kanaan plays for what team? & \multirow{2}{0.26\textwidth}{\{Sporting Al Riyadi Beirut\}} \\
& Paraphrase & What team is Ali Kanaan associated with? &  \\
\midrule
\multirow{4}{*}{Wikidata5m}
& Main Input & Mary Good has relation `award received' to & \multirow{2}{0.26\textwidth}{\{Garvan-Olin Medal; Arkansas Women's Hall of Fame; etc.\}} \\
& Paraphrase & Mary Lowe Good has relation `winner of' to &  \\
\addlinespace[1pt]
& Local Neutral & Mary Good has relation `educated at' to & \{The University of Arkansas; U Arkansas; etc.\} \\
\midrule
\multirow{2}{*}{FEVER}
& Main Input & Tardigrades are also known as space bears. & True \\
& Main Input & The Lion belongs to the genus Vulpes. & False \\
\midrule
\multirow{2}{*}{LeapOfThought}
& Main Input & A viper is a vertebrate. & True \\
& Entailed Data & A viper has a brain. & True \\
\toprule
\end{tabular}
\end{center}
\vspace{-10pt}
\caption{Example datapoint from each dataset, and auxiliary data that accompanies the Main Input. We catalogue examples of noise and other shortcomings for each dataset in Appendix \ref{app:noise-in-datasets}.}
\vspace{-8pt}
\label{tab:data-examples-main}
\end{table*}

\section{Updating Beliefs in Language Models}
\label{sec:updating_beliefs}
Following \citet{de2021editing}, we approach the problem of updating model beliefs as a machine learning problem and train a learned optimizer to perform desired model updates. We discuss metrics for detecting beliefs in Sec. \ref{sec:experiment_detection} and our approach to visualizing belief graphs in Sec. \ref{sec:belief_graphs}.
The core ideas of our approach are outlined in Fig. \ref{fig:main-fig}.

\paragraph{Problem statement and metrics.} We suppose we have a model $f_\theta=p_\theta(y|x)$ parametrized by $\theta$. For an input $x_i$ that has some undesired model output $\hat{y}_i=\argmax_y p_\theta(y|x)$, we wish to obtain a new model $\theta^*$ that produces a desired output $y_i^*$ for $x_i$. 
This new model $\theta^*$ should also fulfill a few other desiderata. As in past work \cite{de2021editing, mitchell2021fast}, we operationalize these desiderata in the following metrics:
\begin{enumerate}
    \item Update Success Rate (\emph{Main Input}): The rate at which the updated model gives the desired output $y_i^*$ for the Main Input $x_i$. 
    \item Update Success Rate (\emph{Paraphrase}): The rate at which the updated model gives the same new prediction for $x_i$ as it does for paraphrases of $x_i$, which are inputs with the same meaning but different surface form. 
    \item Retain Rate (\emph{All Data}): The rate at which the updated model's predictions are unchanged for all other data besides the Main Input.
    \item $\Delta$-Acc (\emph{All Data}): The change in accuracy for the updated model on all other data besides the Main Input.
\end{enumerate}
In practice, Retain Rate (\emph{All Data}) and $\Delta$-Acc are computed with random subsets of a dataset, since these must be computed after every belief update. We add two metrics to those used in past work:  
\begin{enumerate}[start=5]
\item Update Success Rate (\emph{Entailed Data}): The rate at which the updated model makes predictions that are logically entailed by the model's prediction for the Main Input. 
\item Retain Rate (\emph{Local Neutral}): The rate at which the updated model's predictions are unchanged for data that is similar to the Main Input but still logically neutral.
\end{enumerate}
We use Update Success Rate (\emph{Entailed Data}) to measure logical consistency for an updated model, since changing one belief will entail changes in logically entailed beliefs. 
We also split ``retain accuracy" into two cases, one for randomly sampled data as in past work (\emph{All Data}) and the other for specially constructed \emph{Local Neutral} data. Unlike randomly sampled data, Local Neutral data is guaranteed to be logically independent of the Main Input, while still being similar (local) to it.
Together, these six metrics better cover the criteria for belief outlined by \citet{newen2020ascribe}. We compute the metrics using data of the kind shown in Table \ref{tab:data-examples-main}. For a glossary of terms used for these metrics across papers, see Appendix Table \ref{tab:metrics-glossary}.

\paragraph{Evaluation data.} To date, methods have been evaluated on the basis of their ability to change model predictions for all data points, including correctly and incorrectly predicted points. Moreover, the desired labels $\{y_i^*\}_{i=1}^n$ on sequence prediction tasks have each been selected from 
the beam search which produced the original model prediction \cite{de2021editing, mitchell2021fast}. We propose for method evaluation to focus on a more valuable use case: changing the predictions on incorrect points to be correct. In Sec. \ref{sec:experiment_results}, we show that this is a harder task than simply changing predictions to other similar outputs, so the effectiveness of past methods has been overestimated. 

\paragraph{Sequential updates.} The default evaluation procedure in past work on learned optimizers is to update a single model belief, evaluate the new model, then rollback the update before repeating the process for each test point. In Sec. \ref{sec:experiment_results}, we show that it is much harder to update multiple beliefs in a row before evaluating the new model. This is notable because in practice, it is likely that model developers will want to update many beliefs of a trained model, possibly over long timescales, meaning sequential updating is a more realistic application of update methods. We obtain sequential versions of all our metrics by applying $r$ model updates in a row before checking the metrics, meaning there are $\textrm{floor}(n/r)$ measurements for a test set of $n$ points. 

\paragraph{Belief updating method.} We use the \textsc{KnowledgeEditor} architecture from \citet{de2021editing} with our training objective, SLAG. For the details of this architecture, we refer readers to Appendix \ref{app:learned_optimizer_architecture}. Let it suffice for now to observe that a new model is given as a differentiable function
\begin{align*}
    \theta^* = \theta + g_\phi(x_i,\hat{y}_i, y_i^*, \theta)
\end{align*}
using the learned optimizer $g_\phi$, current LM weights $\theta$, Main Input $x_i$, current prediction $\hat{y}_i$, and desired model output $y_i^*$.
In this paper, we generalize the update step to occur in a loop. If we package the above update as $\theta^{(k+1)}=\theta^{(k)}+g_\phi(x_i,\hat{y}_i, y_i^*, \theta^{(k)})$,
then we can obtain new model parameters as 
\begin{align*}
    \theta^* &= \theta^{(k)} + \sum_{j=0}^{K-1} g_\phi(x_i,\hat{y}_i, y_i^*, \theta^{(k+j)})\\
    &= \textrm{Update}(x_i,\hat{y}_i, y_i^*, \theta^{(k)}; \phi, K)
\end{align*}
for a number of steps $K$ from the initial parameters $\theta^{(k)}$. In fact, \citet{de2021editing} use such a loop at test time; we incorporate the loop into training to align the train and test-time distributions.

\paragraph{Learned optimizer training.} The training objective for \textsc{KnowledgeEditor} includes differentiable terms corresponding to Update Success for the Main Input and paraphrases, as well as Retain Rate for all other data. We also include terms for Update Success on entailed data and the Local Neutral Retain Rate, when this is possible given available data. The overall objective requires several kinds of additional data for each point, which we denote by $\mathcal{D}_R$ for other random data, $\mathcal{D}_{LN}$ for local neutral data, $\mathcal{D}_E$ for entailed data, and $\mathcal{D}_P$ for paraphrases of $x_i$. 
For a data point $x_i$ with desired prediction $y_i^*$, the full objective is then:
\begin{align}
    \label{eq:update_objective}
    \begin{split}
    \mathcal{L}&(\phi; x_i,\hat{y}_i, y_i^*, \theta) =
    \lambda_1 \mathcal{L}_{\textrm{Task}}(f_{\theta^*}(x_i), y_i^*) \\
    &+ \lambda_2 \frac{1}{|\mathcal{D}_P|} \sum_{x_P \in \mathcal{D}_P} \mathcal{L}_{\textrm{Task}}(f_{\theta^*}(x_P), y_i^*) \\
    &+ \lambda_3 \frac{1}{|\mathcal{D}_E|} \sum_{x_E,y_E \in \mathcal{D}_E} \mathcal{L}_{\textrm{Task}}(f_{\theta^*}(x_E), y_E) \\
    &+ \lambda_4 \frac{1}{|\mathcal{D}_{LN}|} \sum_{x_{LN} \in \mathcal{D}_{LN}} \textrm{KL}(f_{\theta^*}(x_{LN}) || f_{\theta}(x_{LN})) \\
    &+ \lambda_5 \frac{1}{|\mathcal{D}_R|} \sum_{x_R \in \mathcal{D}_R} \textrm{KL}(f_{\theta^*}(x_R) || f_{\theta}(x_R))
    \end{split}
    \raisetag{25pt}
\end{align}
where $\mathcal{L}_{\textrm{Task}}$ is the loss used to get gradients for $f_\theta$. We use the Cross Entropy loss for binary classification and sequence-to-sequence tasks.

We optimize this objective w.r.t. $\phi$ using AdamW \cite{loshchilov_decoupled_2017}.
To obtain update labels $\{y_i^*\}_{i=1}^n$, we always use the opposite class in binary classification. For sequence-to-sequence tasks, we use the correct label when $\hat{y}_i$ is incorrect, and when $\hat{y}_i$ is correct, we randomly select another label from the training data. 
This choice is in contrast to \citet{de2021editing} and \citet{mitchell2021fast}, who use samples from the model beam search as update labels for all points.

\vspace{2pt}
\noindent\textbf{SLAG objective.} To better prepare the update method for evaluation in a sequential-update setting, we consider training $g_\phi$ to update multiple datapoints in a row. 
Using the per-datapoint loss in Eq. \ref{eq:update_objective}, we obtain our Sequential, Local, and Generalizing (SLAG) loss for a set of $r$ Main Inputs $\mathcal{D}=\{x_i,\hat{y}_i, y_i^*\}_{i=1}^{r}$ as
\begin{align}
    \label{eq:sequential_objective}
    \hspace{-10pt}\mathcal{L}_{\textrm{Sequential}}(\phi; \mathcal{D}, \theta_{t}) \hspace{-2pt} =\hspace{-2pt} \sum_{i=1}^{r} \mathcal{L}(\phi;x_i,\hat{y}_i, y_i^*, \theta_{t+i})
\end{align}
where $\theta_{t+i}=\textrm{Update}(x_i,\hat{y}_i, y_i^*, \theta_{t+i-1}; \phi, K)$ are the model parameters obtained from updating on the first $i$ points in $\mathcal{D}$ (starting from $\theta_t$).
This objective allows us to train $g_\phi$ to update multiple beliefs in a row. To ensure training with this objective is still efficient, we limit how far back through the LM history we backpropagate when computing the gradient w.r.t. $\phi$ for each term in the RHS sum of Eq.~\ref{eq:sequential_objective}. Each parameter vector $\theta_{t}$ depends on $\phi$ and $\theta_{t-1}$. We always apply the stop-gradient function to the most recent vector $\theta_{t-1}$ to prevent backpropagating through it (visualized in Appendix Fig. \ref{fig:objective-loop}). This choice allows our memory use to remain constant in $r$ (see Appendix Fig. \ref{fig:memory_combined}).

\section{Experiment Setup}
\label{sec:experiment_setup}

\subsection{Datasets}

We run experiments with four datasets (example data shown in Appendix Table \ref{tab:data-examples-appendix}). (1) FEVER includes 115,409 True/False factual claims \cite{thorne-etal-2018-fever}. We use the original test set of 10,444 points, and we randomly split the training data into 94,469 train points and 10,496 dev points. (2) zsRE includes 151,631 questions based on relational knowledge from Wikipedia, which we randomly shuffle into train/dev/test splits with 80/10/10\% of the data \cite{levy-etal-2017-zero}. 32.8\% of zsRE questions in each split include paraphrases, and we measure Update Success Rate (\emph{Paraphrase}) for only these points. \citet{talmor2020leap} introduce (3) the LeapOfThought dataset, consisting of 33,484 factual claims that are entailed to be true or false depending on a fact and distractor statements provided as context. 
We drop the distractors from each input and filter the data so that the facts are unique, then shuffle the resulting 14,939 points into train/dev/test splits with 60/10/30\% of the data.

We also construct (4) a sequence prediction task using data from Wikidata5m, which is a relational knowledge base with over 20 million triplets \cite{wang2021kepler}. We build this dataset in order to get Local Neutral data. 
Each input consists of an entity $e_1$ and relation $r$, and the label is another entity $e_2$ that completes the triplet. 
All inputs come in pairs that share the same entity $e_1$ but use different relations with different labels. 
The relations are always one of ten relations that apply to people (see Appendix Table \ref{tab:wikidata-relations}). 
In general, the completion $e_2$ to the Main Input triplet ($e_1$, $r_1$, $e_2$) has no logical consequences for its paired input, ($e_1$, $r_2$, ?). 
This means that changing the model belief for the Main Input should not change its belief for its neutral paired input.
The paired points are also local to the Main Input, i.e. they pertain to the same entity $e_1$ as the Main Input. We obtain four paraphrases for each Main Input using different aliases for the entity and synonyms of the relation. We construct a train set of 150k points and dev and test sets of 10k points each. See Appendix \ref{app:training_details} for further details. 

\begin{table}[t]
\small
\begin{center}
\begin{tabular}{l c c c}
\toprule
& \multicolumn{3}{c}{Belief Consistency $\uparrow$}\\
\cmidrule(lr){2-4}
{Dataset}&{Paraphrase}&{Entailed}&{Contrapos.} \\
    \midrule
    LeapOfThought   &   -  & 85.6 (1.1) &   16.5 (2.7) \\
    zsRE   &   69.5 (1.1)   &   - &   -   \\
    Wikidata5m   &  25.8 (0.5)   &   - &   -     \\
    \bottomrule
\end{tabular}
\end{center}
\vspace{-10pt}
\caption{Belief metric results across datasets.}
\vspace{-6pt}
\label{tab:belief_metrics}
\end{table}

\subsection{Methods Evaluated}

\begin{table}[t]
\small
\begin{center}
\begin{tabular}{l c c}
\toprule
& \multicolumn{2}{c}{Paraphrase Consistency $\uparrow$}\\
\cmidrule(lr){2-3}
{Dataset}&{Model Incorrect}&{Model Correct} \\
    \midrule
    zsRE   &   61.39 (1.33) &   91.82 (1.17)    \\
    Wikidata5m   &   24.55 (0.48) &   37.20 (2.06)   \\
    \bottomrule
\end{tabular}
\end{center}
\vspace{-9pt}
\caption{Paraphrase consistency by the correctness of the model prediction on the Main Input.}
\vspace{-10pt}
\label{tab:belief_metrics_by_correctness}
\end{table}

\noindent\textbf{Models.} 
We train five models with different random seeds for each dataset, using RoBERTa-base for binary tasks and BART-base for sequence-to-sequence tasks (accuracies in Appendix Table \ref{tab:belief_metrics_appendix}).
For each of the five models, we train one learned optimizer using SLAG and one with the objective from \citet{de2021editing}, which we list as KE in tables below.
Our model selection criterion is the mean of: the average Update Success Rate (across data types), Retain Rate (only for Local Neutral data), and $\Delta$-Acc for All Data. 
We tune the choice of SLAG objective terms for each task separately (see Appendix Table \ref{tab:method-hyperparameters} for final selections; results discussed in Sec. \ref{sec:experiment_objective}).
Other hyperparameters are given in Appendix \ref{app:training_details} and memory use is shown in Appendix Fig. \ref{fig:memory_combined}. 
To summarize the differences between SLAG and \textsc{KnowledgeEditor}: (1) we use $K_{\textrm{train}}=K_{\textrm{test}}$ rather than $K_{\textrm{train}}=1$; (2) we adopt training labels using real data labels rather than alternatives from the model's beam search; and (3) our objective terms differ following tuning.

\vspace{2pt}
\noindent\textbf{Baselines.} We use off-the-shelf optimizers as baselines. We tune the baseline hyperparameters separately for each dataset, selecting among several kinds of optimizers, learning rates, and the number of update steps. The selection criterion is the same as the criterion outlined for learned optimizers above. The resulting baselines are surprisingly strong (see Appendix Table \ref{tab:baseline-hyperparameters} for final selections). 

\vspace{2pt}
\noindent\textbf{Hypothesis testing.} We obtain 95\% confidence intervals and perform hypothesis tests via block bootstrap, resampling model seeds and data points \cite{efron1994introduction}. For ablation experiments, we run only one model seed per condition.

\begin{table*}[t]
\small
\begin{center}
\begin{tabular}{l l r c c c r r}
\toprule
\multicolumn{2}{l}{\textbf{Single-Update Setting}}   & \multicolumn{3}{c}{Update Success Rate}  & \multicolumn{2}{c}{Retain Rate} & \multicolumn{1}{c}{$\Delta$-Acc} \\
\cmidrule(lr){3-5} \cmidrule(lr){6-7} \cmidrule(lr){8-8}
Dataset & Method & \multicolumn{1}{c}{Main Input} & \multicolumn{1}{c}{Paraphrases} & \multicolumn{1}{c}{Entailed Data} & \multicolumn{1}{c}{Local Neutral}         & \multicolumn{1}{c}{All Data}  & \multicolumn{1}{c}{All Data}  \\
\midrule
\multirow{3}{*}{FEVER}           & AdamW     & 100 (0.0) & -           & -             & -                    & \textbf{98.80} (0.2)            & \textbf{0.22} (0.1)      \\ 
           & KE     & 99.98 (<0.1) & -           & -             & -                    & 98.28 (0.3)            & -0.24 (0.1)      \\
           & SLAG     & 99.99 (<0.1) & -           & -             & -                    & 98.41 (0.2)            & -0.20 (0.1)      \\
\midrule
\multirow{3}{*}{LeapOfThought} & SGD & 100 (0.0)       & -           & 72.48 (4.6)          & -                    & 95.52 (0.4)            & \textbf{1.23} (0.8)      \\
 & KE & 99.78 (0.4)       & -           & 74.48 (4.4)          & -                    & 93.50 (1.3)        & -1.33 (1.1)      \\
 & SLAG & 100 (0.0)       & -           & 75.50 (4.3)          & -                    & 94.92 (1.4)            & -1.31 (1.2)      \\
\midrule
\multirow{3}{*}{zsRE}            & SGD     & \textbf{99.36} (0.1)       & 94.44 (0.6)        & -             & -                    & 74.73 (0.4)           & -0.43 (0.1) \\
            & KE     & 84.73 (1.4)      & 89.26 (1.8)        & -             & -                    & 71.55 (2.4)            & -2.19 (0.4) \\
            & SLAG     & 94.29 (0.4)       & 94.71 (0.5)        & -             & -                    & \textbf{80.48} (1.3)    & -0.29 (0.1)      \\
\midrule
\multirow{3}{*}{Wikidata5m}      & SGD     & \textbf{98.05} (0.3)      & 68.78 (0.8)        & -             & 41.46 (1.0)                 & 58.62 (0.6)            & -1.97 (0.3)   \\  
      & KE     & 74.57 (2.9)      & 58.05 (2.2)        & -             & 40.84 (1.8)                 & 53.58 (2.2)            & -3.03 (0.5)   \\  
      & SLAG     & 87.59 (0.6)      & \textbf{80.70} (0.9)        & -             & \textbf{47.85} (1.0)                 & \textbf{63.51} (1.3)            & -1.71 (0.3)   \\ 
\bottomrule
\end{tabular}
\end{center}
\vspace{-10pt}
\caption{Belief update metrics for off-the-shelf optimizers, \textsc{KnowledgeEditor} (KE) from \citet{de2021editing}, and SLAG, with $r_{\textrm{test}}=1$. Bolded numbers are the best in their group at a statistical significance threshold of $p<.05$ (or lower). Our SLAG objective improves over KE, but off-the-shelf optimizers perform surprisingly well.}
\vspace{-8pt}
\label{tab:update_table}
\end{table*}

\section{Experiment Results}
\label{sec:experiment_results}

\subsection{Do LMs have beliefs about the world?} 
\label{sec:experiment_detection}
We measure Paraphrase Consistency, Entailment Acc, and Contrapositive Acc for our finetuned task models. Paraphrase Consistency is the fraction of paraphrase pairs for which a model produces the same output \cite{elazar2021measuring}. Entailment Acc is the accuracy of a model on data that is entailed by the Main Input.
For LeapOfThought (see Table \ref{tab:data-examples-main}), ``Main Input $x_i$ is true" implies ``entailed input $x_E$ has label $y_E$," but the inverse ($\neg A \Rightarrow \neg B$) does not necessarily hold. Therefore, we compute Entailment Acc only where the Main Input prediction is correct.
We do know that the contrapositive holds: ``Entailed input $x_E$ does not have label $y_E$" implies that ``Main Input $x_i$ is false." So for Contrapositive Acc, we measure how often the model follows this rule, when the antecedent holds of its prediction.

\vspace{2pt}
\noindent\textbf{Belief measurement results.} Table \ref{tab:belief_metrics} shows the belief metrics for each dataset.
We find that ${\small \sim}$100M parameter models show limited evidence of having beliefs about the world. Paraphrase consistency is 69.50\% ($\pm$ 1.09) for zsRE and much lower for Wikidata5m (25.84\%$\pm$0.53). While entailment accuracy is high for LeapOfThought (85.63\%$\pm$1.08), the model is consistent under the contrapositive only 16.51\% ($\pm$ 2.71) of the time. One might reasonably set the bar for qualifying as a ``belief" higher than these scores. But since belief-likeness comes in degrees, we continue to refer to model beliefs for the rest of the paper. Interestingly, the metrics are much higher when the model prediction on the Main Input is correct (Table \ref{tab:belief_metrics_by_correctness}).

\subsection{Can we update beliefs in LMs?} 
\label{sec:experiment_updating}
First, we compare two evaluation procedures for sequence prediction tasks: correcting model beliefs versus changing them to an alternative from the model's beam search. We do so for zsRE using SLAG. 
Next, we compare belief update metrics across datasets using \textsc{KnowledgeEditor}, SLAG, and off-the-shelf optimizers as baselines.
We report results in single-update ($r_\textrm{test}=1$) and sequential-update ($r_\textrm{test}=10$) settings. See Appendix Fig. \ref{fig:r_ablation} for an ablation across $r_\textrm{test}$.

\begin{table}[t]
\small
\begin{center}
\begin{tabular}{l c c c}
\toprule
& \multicolumn{2}{c}{Update Success Rate $\uparrow$}  & $\Delta$-Acc $\uparrow$ \\
\cmidrule(lr){2-3} \cmidrule(lr){4-4}
Desired Label & Main Input & Paraphrase & All Data  \\
\midrule
Beam Label & 97.41 (0.3) & 97.03 (0.4) & -0.30 (0.1) \\
Correct Label & 94.46 (0.7) & 94.45 (0.7) & -0.24 (0.1) \\
\bottomrule
\end{tabular}
\end{center}
\vspace{-10pt}
\caption{
Evaluation difficulty by desired model output, for a learned optimizer trained with SLAG on zsRE.
}
\vspace{-11pt}
\label{tab:eval_ablation}
\end{table}

\begin{table*}[t]
\small
\begin{center}
\begin{tabular}{l l r c c r r r}
\toprule
\multicolumn{2}{l}{\textbf{Sequential-Update Setting}}  & \multicolumn{3}{c}{Update Success Rate}  & \multicolumn{2}{c}{Retain Rate} & \multicolumn{1}{c}{$\Delta$-Acc} \\
\cmidrule(lr){3-5} \cmidrule(lr){6-7} \cmidrule(lr){8-8}
Dataset & Method & \multicolumn{1}{c}{Main Input} & \multicolumn{1}{c}{Paraphrases} & \multicolumn{1}{c}{Entailed Data} & \multicolumn{1}{c}{Local Neutral}         & \multicolumn{1}{c}{All Data}  & \multicolumn{1}{c}{All Data}  \\
\midrule
\multirow{3}{*}{FEVER}           & AdamW     & 92.81 (1.3) &\multicolumn{1}{c}{-}          &\multicolumn{1}{c}{-}            &\multicolumn{1}{c}{-}                   & \textbf{91.86} (1.4)            & \textbf{1.16} (0.6)      \\
& SLAG$_1$     & 74.13 (1.8) &\multicolumn{1}{c}{-}          &\multicolumn{1}{c}{-}            &\multicolumn{1}{c}{-}                   & 39.86 (0.7)            & -27.13 (1.3)      \\
& SLAG$_{10}$     & 91.27 (2.9) &\multicolumn{1}{c}{-}          &\multicolumn{1}{c}{-}            &\multicolumn{1}{c}{-}                   & 70.30 (5.8)            & -11.96 (4.5)      \\
\midrule
\multirow{3}{*}{LeapOfThought}
& SGD & 100 (0.0) &\multicolumn{1}{c}{-}& \textbf{61.34} (5.0) &\multicolumn{1}{c}{-}& \textbf{82.62} (0.8) & \textbf{-4.93} (1.0) \\
& SLAG$_1$ & 96.14 (2.3) &\multicolumn{1}{c}{-}& 49.27 (6.0) &\multicolumn{1}{c}{-}& 72.45 (0.9) & -15.03 (1.0) \\
& SLAG$_{10}$ & 100 (0.0) &\multicolumn{1}{c}{-}& 50.46 (5.5) &\multicolumn{1}{c}{-}& 74.02 (1.1) & -13.03 (1.3) \\
\midrule
\multirow{3}{*}{zsRE}            
& SGD & 82.71 (0.6) & 90.81 (0.7) &\multicolumn{1}{c}{-}&\multicolumn{1}{c}{-}& 40.49 (0.6) & -2.38 (0.3) \\
& SLAG$_1$ & 0.10 (<0.1) & 36.55 (1.4) &\multicolumn{1}{c}{-}&\multicolumn{1}{c}{-}& 0.05 (<0.1) & -20.98 (0.7) \\
& SLAG$_{10}$ & \textbf{87.57} (0.6) & \textbf{92.20} (0.7)   &\multicolumn{1}{c}{-}&\multicolumn{1}{c}{-}& \textbf{47.19} (0.7) & \textbf{-1.74} (0.3) \\
\midrule
\multirow{3}{*}{Wikidata5m}
& SGD & 56.82 (0.8) & 54.49 (0.7) &\multicolumn{1}{c}{-}& 6.40 (0.4) & 26.37 (0.6) & -3.96 (0.4) \\
& SLAG$_1$ & 0 (0.0) & 40.84 (0.9) &\multicolumn{1}{c}{-}& 0 (0.0) & 0 (0.0) & -10.05 (0.6) \\
& SLAG$_{10}$ & \textbf{58.27} (1.0) & \textbf{65.51} (0.9) &\multicolumn{1}{c}{-}& \textbf{7.36} (0.5) & \textbf{27.76} (0.7) & -3.62 (0.4) \\
\bottomrule
\end{tabular}
\end{center}
\vspace{-10pt}
\caption{Belief update results when a model is sequentially updated $r_{\textrm{test}}{=}10$ times. SLAG$_R$ uses $r_{\textrm{train}}{=}R$. On sequence prediction tasks in this setting, SLAG can outperform the off-the-shelf optimizers across metrics.}
\vspace{-9pt}
\label{tab:sequential_update_table}
\end{table*}

\vspace{2pt}
\noindent\textbf{Correcting beliefs vs. changing beliefs.} Given the results in Table \ref{tab:eval_ablation}, we find that correcting model outputs is harder than simply changing them to a plausible alternative. Update Success can rise by a full 2.96 ($\pm$0.48; $p{<}$\num{1e-4}) points for Main Inputs and 2.58 ($\pm$0.81; $p{<}$\num{1e-4}) for Paraphrases, while $\Delta$-Acc is virtually unchanged. 
This suggests that that past work has overestimated the efficacy of belief update methods for actually fixing models. Henceforth we evaluate methods according to their ability to update model beliefs to be true.

\vspace{2pt}
\noindent\textbf{Update method results (single update).} Table \ref{tab:update_table} shows the results in a single-update setting. First, we find that off-the-shelf optimizers are very effective across the board. The baselines show Main Input Update Success Rates of 100\% for binary tasks with positive $\Delta$-Acc scores.\footnote{Positive $\Delta$-Acc values are possibly due to distribution shift in the test split. In FEVER, for instance, the train and dev data are 73\% True, while test data is 50\% True. On the dev split, AdamW achieves a negative $\Delta$-Acc, -0.18 ($\pm$0.11).}
On sequence prediction tasks, SGD achieves 98\%+ Main Input Update Success with competitive $\Delta$-Acc scores.
When strongly tuned, these baselines outperform learned optimizers on most metrics here.

However, SLAG does surpass the baselines in a few places. All Data Retain Rate on zsRE rises by 5.77 points ($\pm$1.43; $p{<}$\num{1e-4}), and on Wikidata5m we improve Paraphrase Update Success by 11.92 points ($\pm$1.20; $p{<}\num{1e-4}$) and the Local Neutral Retain Rate by 6.40 ($\pm$1.41; $p{<}$\num{1e-4}) points. 
The gain on Entailed Data Update Success is 3.02 points, but it is not significant ($\pm$6.26; $p{=}$.345). The SLAG objective also greatly improves performance over KE for sequence prediction tasks. 

\vspace{2pt}
\noindent\textbf{Update method results (sequential updates).} We give results for a sequential update setting ($r_\textrm{test}{=}10$) in
Table \ref{tab:sequential_update_table}. 
Immediately we see this is a much more difficult setting for updating model beliefs, as update metrics are generally much lower for each dataset. 
Next, we observe that learned optimizers with SLAG$_{10}$ ($r_\textrm{train}{=}10$) now outperform baselines on sequence prediction tasks. 
On zsRE, we improve Update Success for Main Inputs by 4.86 ($\pm$0.83; $p{=}$\num{1e-4}) and for Paraphrases by 1.39 ($\pm$0.93; $p{=}$.004), with better $\Delta$-Acc by 0.64 ($\pm$0.35; $p{=}$.0005). Improvements trend in the same direction for Wikidata5m and are all statistically significant except for the gain in $\Delta$-Acc. The jump on Paraphrases in particular is very large (11.02$\pm$1.17; $p{<}$\num{1e-4}). In comparison, using a non-sequential ($r_{\textrm{train}}=1$) training objective leads to drastic drops in performance.

Learned optimizers still struggle with the binary datasets compared to the off-the-shelf optimizers. The baselines achieve high update update success with much better $\Delta$-Acc scores, by 13.12 ($\pm$4.51; $p{=}$\num{1e-4}) on FEVER and 8.16 ($\pm$1.63; $p{=}$\num{1e-4}) on LeapOfThought. Also on LeapOfThought, the baseline's update success with entailed data is over 10 points higher ($\pm$7.38; $p{=}$.004). 

\subsection{How does the learned optimizer objective influence performance?} 
\label{sec:experiment_objective}
Here, we discuss ablations with respect to the terms in the training objective, Eq. \ref{eq:update_objective}. We show the effect of $K_{\textrm{train}}$ in Appendix Fig. \ref{fig:k_ablation} and the choice of optimizer training labels in Appendix Table \ref{tab:label_ablation}.

\vspace{2pt}
\noindent\textbf{Training objective ablation.} We give objective ablation results in Appendix Table \ref{tab:objective_term_ablation}.
Surprisingly, we do not always see that the objective terms help for the data they are intended to help with. First, we obtain mixed results for the paraphrase objective. On zsRE, the objective term seems to hinder performance, with update success dropping on Main Inputs by 0.71 ($\pm$0.60; $p{=}$.021) and $\Delta$-Acc dropping by 0.18 ($\pm$0.19; $p{=}$.069), while the paraphrase Update Success Rate itself is unaffected. With Wikidata5m, however, the paraphrase term improves paraphrase update success by a large margin of 16.94 ($\pm$1.03; $p{<}$\num{1e-4}) points.
Adding the Local Neutral (LN) term with the paraphrase term greatly improves the LN Retain Rate for Wikidata5m, by 9.71 points ($\pm$1.44; $p{<}$\num{1e-4}), though both of these terms come at a cost to Main Input Update Success, similar to zsRE. 
Lastly, we do not find that the entailment objective improves Entailed Data Update Success; in fact, this metric falls by 4.56 ($\pm$7.22; $p{=}.213$) points with the objective. 

\section{Analysis}

\begin{table}[t]
\small
\begin{center}
\begin{tabular}{l c c }
\toprule
{Metric}&{Before Update}&{After Update} \\
    \midrule
    Entailment Acc   &  \hspace{3pt} 58.30 (5.7)*   &  75.50 (4.3)      \\
    Para. Cons (zsRE)   & 61.39 (1.3) &   94.53 (0.6)   \\
    Para. Cons (Wiki)  &   24.69 (0.5)  &      84.56 (0.9)   \\
    \bottomrule
\end{tabular}
\end{center}
\vspace{-11pt}
\caption{
Entailment Acc and Paraphrase Consistency before and after model updates to incorrect points.
*All Main Inputs in this subset are wrongly predicted as false, so the entailment does not actually hold.
}
\vspace{-10pt}
\label{tab:update_metrics_before_after_update}
\end{table}

\subsection{Belief updates improve consistency} 
\begin{figure}[t]
    \centering
    \includegraphics[width=0.4\textwidth]{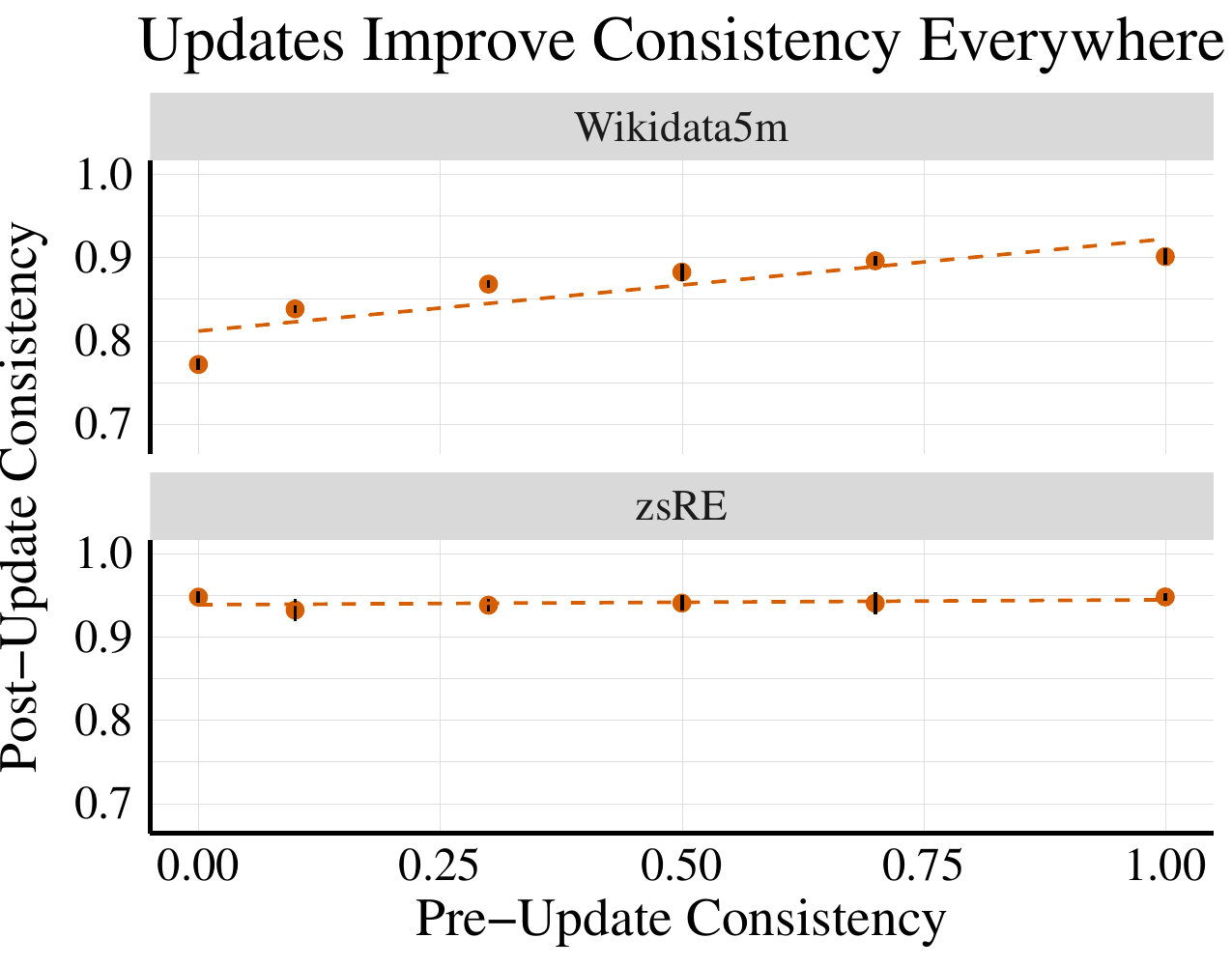}
    \vspace{-8pt}
    \caption{
    Post-update consistency under paraphrase is high even for points with low pre-update consistency.
    }
    \label{fig:after-cons-by-before-cons}
    \vspace{-11pt}
\end{figure}

In Table \ref{tab:update_metrics_before_after_update}, we show belief metrics before and after model updates using SLAG with $r_{\textrm{test}}{=}1$. We observe that belief updates greatly improve paraphrase consistency and entailment accuracy for updated data. Paraphrase consistency rises by 
33.14$\pm$1.46 on zsRE and 59.87$\pm$1.09 on Wikidata5m, while Entailment Acc rises by 17.20$\pm$7.10 points.
To see if these improvements depend on pre-update consistency, we plot paraphrase consistency before and after updating in Fig. \ref{fig:after-cons-by-before-cons}.
For zsRE, consistency rises irrespective of pre-update consistency. 
There is a noticeable trend for Wikidata5m paraphrases, where post-update consistency is 90.1\% when pre-update consistency is maxed out, versus 77.1\% for totally inconsistent pre-update beliefs.
We conclude that learned optimizers can induce a fairly consistent model belief even where there is no consistent belief to begin with.

\subsection{Which beliefs are hard to retain when updating other beliefs?}
\label{sec:which_beliefs_are_sensitive}
We find that the Retain Rate depends heavily on whether the predictions on that data are correct to begin with.
On zsRE for instance, the retain rate on correct inputs is about 96\%, while for incorrect predictions, it is about 75\%. 
So it appears that incorrect predictions are the most sensitive to model updates, and these points merely change from one incorrect prediction to another. On FEVER, incorrect beliefs change around 4\% of the time, while correct beliefs change only 2.5\% of the time.

We also find that Local Neutral beliefs are much harder to avoid changing than simply random data. For Wikidata5m in Table \ref{tab:update_table}, the Retain Rate on All Data is 61.51$\pm$1.33, while for Local Neutral data it is a full 15.66 points lower, at 47.85$\pm$0.96. 

\subsection{Belief Graphs} 
\label{sec:belief_graphs}

We now construct \emph{belief graphs} for the purpose of better understanding the connections between model beliefs. 
We form the graphs from a set of datapoints by updating each prediction and checking what other predictions change.
We represent each datapoint as its own node in a belief graph. Whenever updating a datapoint $u$ changes the model prediction for point $v$, we draw a directed edge from $u$ to $v$.
Following our results in Sec. \ref{sec:experiment_updating}, we use off-the-shelf optimizers to change the model output to the opposite of its original prediction for every datapoint.
The resulting graphs have up to $n^2-n$ edges (no self edges). For FEVER we obtain a graph of 10,444 nodes, and for LeapOfThought we obtain a graph with 8642 nodes, which is double the original test set size because we include both Main Inputs and Entailed Data as their own nodes. 

\begin{figure*}[t]
    \centering
    \includegraphics[width=0.88\textwidth]{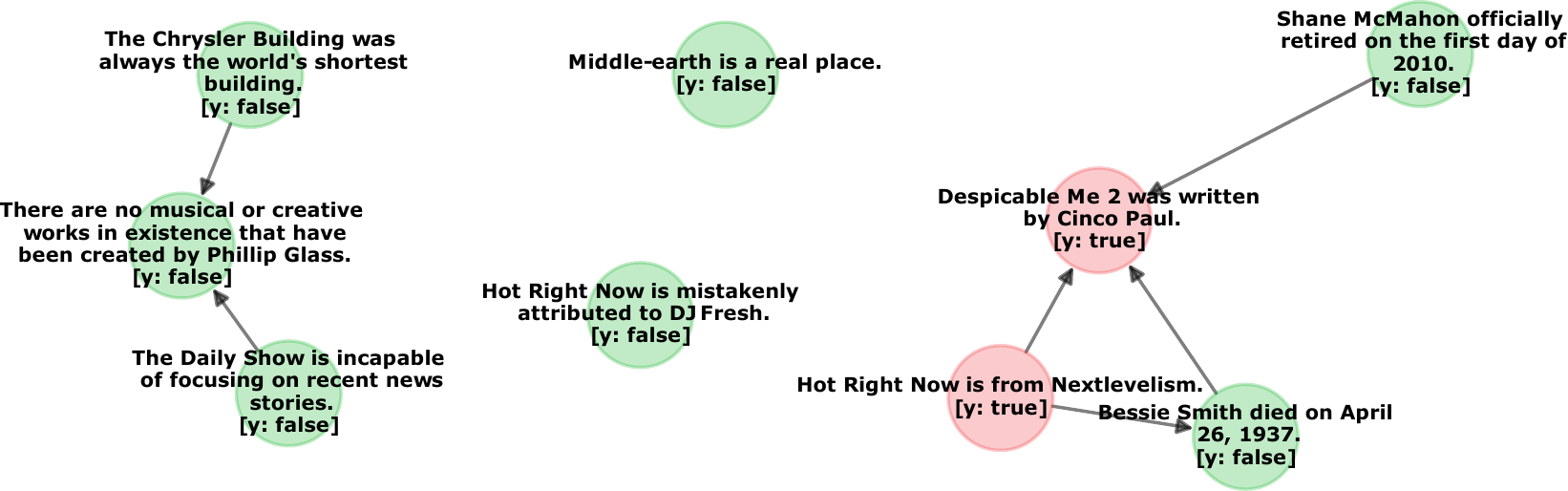}
    \caption{A non-random subgraph of the belief graph for a model trained on FEVER. Directed edges from $u$ to $v$ indicate that changing the model belief in $u$ causes the belief in $v$ to change. The ground-truth label is given in brackets for each point, and node color shows the model's accuracy before any updates (green=correct).}
    \label{fig:belief-graph}
    \vspace{-8pt}
\end{figure*}

We visualize part of a belief graph in Fig. \ref{fig:belief-graph}. This figure shows a non-random subgraph intended to give a representative view of the data (we give three random subgraphs of 20 nodes in Appendix \ref{app:additional-results}). On inspection, we see no reason that beliefs are connected or not connected. 
Whether or not changing one belief changes another appears essentially random.
We come to same conclusion looking at other random subgraphs (see Appendix Figures \ref{fig:random-subgraph-0}, \ref{fig:random-subgraph-1}, \ref{fig:random-subgraph-2}).
However, we do know of some aggregate trends from earlier results. Sec. \ref{sec:which_beliefs_are_sensitive} suggests that a model's incorrect beliefs are most likely to change after model updates, and following Sec. \ref{sec:experiment_results}, we have reason to believe that Local beliefs are more likely than others to change with model updates.

We highlight a few summary statistics here from Table \ref{tab:graph-statistics} for a broader view of the graphs. 
First, \%~Edgeless is the proportion of nodes which have no in or out edges. Since this is 0 for both datasets, every belief can be changed by editing the right belief. 
\#~In Edges is the number of in edges at the 95\textsuperscript{th} percentile, meaning 5\% of beliefs have more in edges than this value, and the same holds of \#~Out Edges. These values grow to a rather large fraction of the overall datasets, suggesting that (1) some beliefs are sensitive to changes in a large fraction of all beliefs, and (2) some beliefs are influential to hundreds of other beliefs when changed. 
\#~Corrupted is the number of correct predictions changed to be incorrect following a model update. For 5\% of the data, model updates cause at least 211 points to become incorrectly predicted on FEVER, and 2,752 points for LeapOfThought.
Lastly, \%~Update-Transitivity represents the answer to the question: if updating belief A changes belief B, and updating belief B changes belief C, what proportion of the time does updating A change C? 
For these datasets, a logically consistent model should display 100\% Update-Transitivity (see Appendix \ref{app:metric_and_bootstrap_details} for a caveat on this metric). We find that belief updates often yield intransitive results for both datasets.

\begin{table}[t]
\begin{center}
\small
\begin{tabular}{l c c}
\toprule 
& \multicolumn{2}{c}{Dataset} \\
\cmidrule(lr){2-3}
Metric & FEVER & LeapOfThought \\
\midrule
\# Nodes & 10,444 & 8,642\\
\% Edgeless & 0.0 & 0.0\\
\# Edges Total & 1.88m & 9.71m\\
\# In Edges  (95\textsuperscript{th} perc.) & 1,088 & 5,347\\
\# Out Edges  (95\textsuperscript{th} perc.) & 390 & 3,087\\
\# Corrupted (95\textsuperscript{th} perc.)& 211 & 2,752\\
\% Update-Transitivity & 66.64 & \hspace{2pt} 24.38*\\
\bottomrule
\end{tabular}
\end{center}
\vspace{-10pt}
\caption{Belief graph summary statistics. *We compute Update-Transitivity for LeapOfThought with $n=4000$ points due to computational cost.}
\vspace{-10pt}
\label{tab:graph-statistics}
\end{table}

\section{Discussion and Conclusion}

\vspace{2pt}
\noindent\textbf{Degrees of commitment to beliefs.} The data we use comes in the form of declarative statements and answers to questions. These utterances take what is called a veridical stance toward a proposition, displaying a ``full commitment" to that proposition's truthfulness \cite{giannakidou2020linguistic}. It will be valuable for future work to explore two dimensions of uncertainty in beliefs: (1) expression of uncertainty in language, via partial or trivial commitments (like ``X might be Y") and (2) expression of uncertainty mathematically, via probabilities assigned by a model to utterances or True/False values. In this paper we treat a belief as ``updated" when the model output changes, but this ignores any underlying change in the distribution $p_\theta(y|x)$ that could occur even if its mode does not change.

\vspace{2pt}
\noindent\textbf{Ethics and dual use concerns.} 
Belief update methods may be used to either correct undesired beliefs or induce problematic beliefs in LMs, and it is not clear whether these capabilities could be separated.
We propose to evaluate methods only on the basis of their ability to correct mistaken model beliefs, but the malicious use case remains. 
We are uncertain about how a bad belief would influence the general behavior of a model (e.g. answers to many questions), but it is possible that a belief update method could instill bad beliefs in a generally capable LM with far-reaching implications for model behavior.
That said, we hope that these methods will instead be used to update undesirable moral, social, and factual beliefs in large LMs.

\vspace{2pt}
\noindent\textbf{Conclusion.} We first discuss criteria for detecting when LMs have \emph{beliefs} about the world. 
Next, we argue for evaluating belief update methods by their ability to correct mistaken beliefs, which is harder than the evaluation done in past work.
We show that strongly tuned off-the-shelf optimizers make for surprisingly good belief update methods, even surpassing specialized learned optimizers in several settings. But with a new training objective (SLAG), we are able to outperform these baselines on sequence prediction tasks when updating multiple beliefs one after another.
Finally, we introduce \emph{belief graphs} as a means of understanding the connections between model beliefs. We find that model beliefs are highly interconnected, with some beliefs influencing hundreds of other beliefs. While it is hard to point to concrete reasons for individual connections between beliefs, we identify several patterns in the dependencies between beliefs.

\bibliography{main}
\bibliographystyle{acl_natbib}

\appendix

\begin{figure*}[t]
    \centering
    \includegraphics[width=0.8\textwidth]{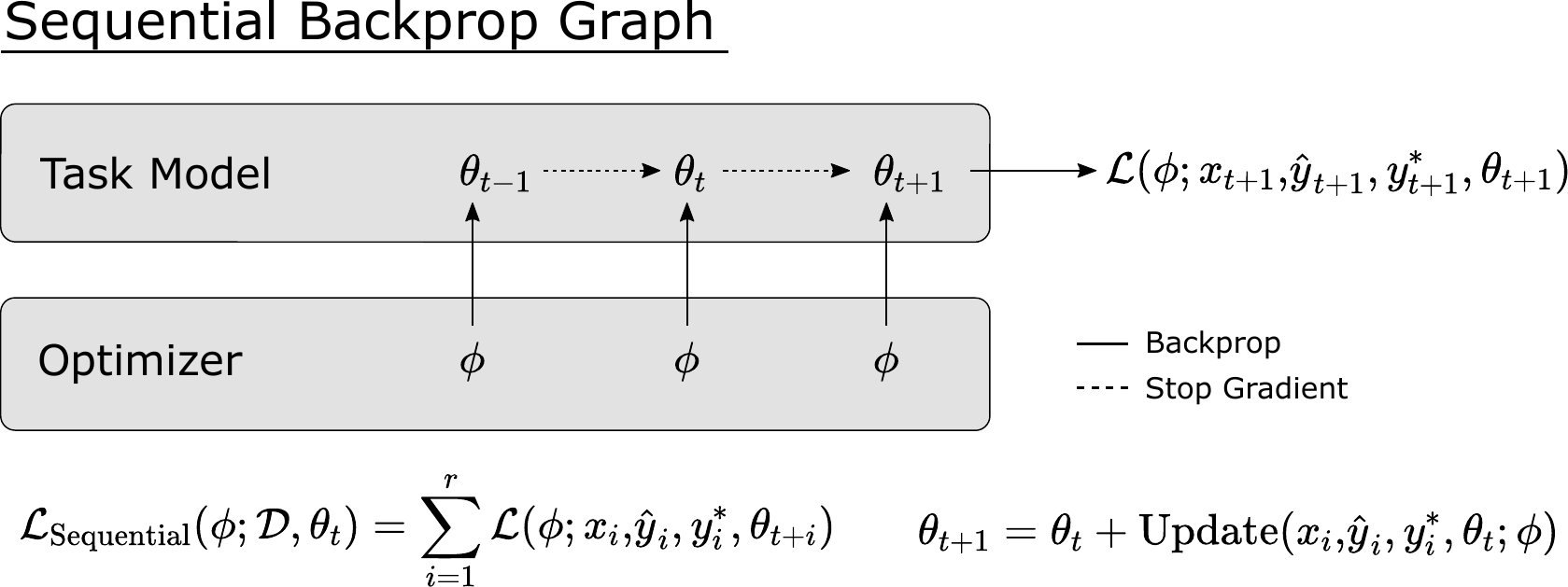}
    \caption{The backpropagation graph for sequential model updates.}
    \label{fig:objective-loop}
    \vspace{-8pt}
\end{figure*}

\section{Learned Optimizer Details}
\label{app:learned_optimizer_architecture}

\vspace{2pt}
\noindent\textbf{Architecture.}
\textsc{KnowledgeEditor} is a learned optimizer $g: \mathcal{X} \times \mathcal{Y} \times \mathcal{Y} \times  \Theta \rightarrow \Theta$ that produces new model weights by applying an adjusted gradient step to a model. For reference, we give a glossary of symbols used here in Table \ref{tab:symbols-glossary}. For additional details beyond what is presented here, we refer readers to \citet{de2021editing}. 

At a high level, $g_\phi$ first encodes an input $x_i$ and requested prediction change into a vector $h$, then processes $h$ into two low-rank matrices $A$ and $B$ that are used to transform the model gradient on $x_i$, $\nabla_\theta\mathcal{L}(x_i,y_i^*)$. 
For Transformer models, the method edits only attention and feed-forward weights, so all model gradients match the shape of an associated weight matrix of shape $d_1 \times d_2$.
Formally, a new model $\theta^*$ is obtained using a learned optimizer $g_\phi$ as follows:
\begin{align*}
    h &= \textrm{LSTM}([x; \hat{y}; y^*]) \\
    \{u,v,\gamma,\delta\} &= \{\textrm{MLP}_i(h)\}_{i=1}^4 \\
    A &= \textrm{softmax}(u)v^T \\
    B &= \textrm{softmax}(\gamma)\delta^T \\
    \eta &= \sigma(\textrm{MLP}(h))\\
    \theta^* &= \theta + \eta(A \circ \nabla_\theta\mathcal{L}(x_i,y_i^*) + B)
\end{align*}
where $\phi$ consists of all LSTM and MLP parameters.

\vspace{2pt}
\noindent\textbf{Training Algorithm.} The learned optimizer objective is optimized w.r.t. $\phi$ with AdamW through a standard procedure of randomly sampling minibatches without replacement \cite{loshchilov_decoupled_2017}. Within each batch, one datapoint is randomly selected as the Main Input, and the remaining points are used as $\mathcal{D}_R$. 
To obtain update labels $\{y_i^*\}_{i=1}^n$, we always use the opposite class in binary classification. For sequence-to-sequence tasks, we use the correct label when $\hat{y}_i$ is incorrect, and when $\hat{y}_i$ is correct, we randomly select another label from the training data. 
This choice is in contrast to \citet{de2021editing} and \citet{mitchell2021fast}, who use samples from the model beam search as update labels for all points.

\begin{table}
    \centering
    \small
    \begin{tabular}{p{.15\textwidth} p{.24\textwidth}}
    \toprule
    \multicolumn{2}{c}{Symbol Glossary}\\
    \midrule
         $f_\theta$ & Language Model \\
         \addlinespace[1pt]
         $g_\phi$ & Learned optimizer \\
         \addlinespace[1pt]
         $x_i$ & Main Input \\
         \addlinespace[1pt]
         $\hat{y}_i$ & LM output on $x_i$ \\
         $y_i^*$ & Desired output \\
         \addlinespace[2pt]
         $\nabla_\theta\mathcal{L}(x_i,y_i^*)$ & Gradient of LM \\
         $\textrm{Update}(x_i,\hat{y}_i, y_i^*, \theta)$ & Update one LM belief \\
         \addlinespace[2pt]
         $\mathcal{L}(\phi; x_i,\hat{y}_i, y_i^*, \theta)$ & Belief update objective for $x_i$\\
         \addlinespace[2pt]
         $\mathcal{L}_{\textrm{Sequential}}(\phi; \mathcal{D}, \theta_{t})$ & Sequential objective (SLAG) \\
         \addlinespace[1pt]
         $K$ & \# gradient steps in Update($\cdot$) \\
         $r$ & \# beliefs updated in $\mathcal{L}_{\textrm{Sequential}}$ \\
     \bottomrule
    \end{tabular}
    \vspace{-3pt}
    \caption{Symbol descriptions for the learned optimizer.}
    \vspace{-10pt}
    \label{tab:symbols-glossary}
\end{table}

\section{Additional Training Details}
\label{app:training_details}

\subsection{Compute Costs.}
\label{app:compute-costs}

\begin{figure*}
    \centering
    \includegraphics[width=0.89\textwidth]{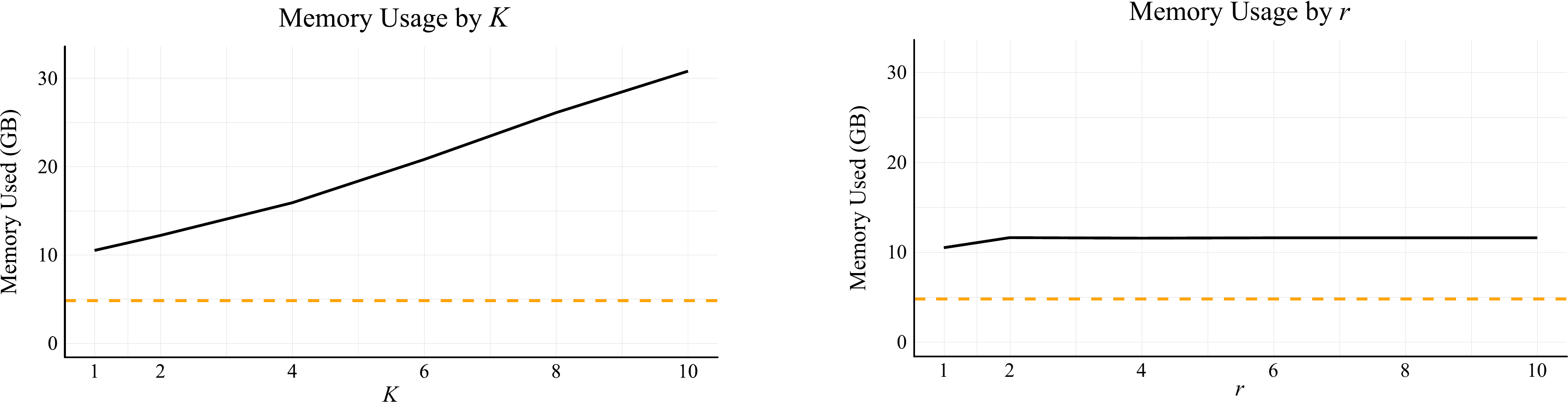}
    \vspace{-10pt}
    \caption{Training memory usage in terms of $K$ and $r$ hyperparameters in our implementation, for a learned optimizer trained for a BART-base model on zsRE, using a batch size of 16. For comparison, the orange dashed line shows the memory use of training the BART-base model on zsRE, using the same batch size. Our use of the stop-gradient function limits the growth of runtime and memory w.r.t. both $K$ and $r$. By accumulating gradients across points, memory w.r.t. $r$ is kept constant. The same trick could be applied to the $K$ looped gradient steps inside the Update function, at the trade-off of backpropagating $K$ times per point rather than one time.}
    \label{fig:memory_combined}
    \vspace{-5pt}
\end{figure*}

\vspace{2pt}
\noindent\textbf{Learned optimizer memory.} 
The hypernetwork has 92m trainable parameters for RoBERTa-base (which is 125m parameters), and 105m parameters for BART-base (which is 139m parameters). 
To increase training efficiency, we limit how far into the task model history we backpropagate. As shown in Fig. \ref{fig:objective-loop}, when backpropagating through task model parameters $\theta_t=\theta_{t-1}+\textrm{Update}(x_i,\hat{y}_i, y_i^*, \theta_{t-1}; \phi)$, we continue backpropagating through $\textrm{Update}(x_i,\hat{y}_i, y_i^*, \theta_{t-1})$ but \emph{not} $\theta_{t-1}$, which is also dependent on $\phi$. That is, we apply a stop-gradient function to $\theta_{t-1}$. This way, we compute the derivative $\nabla_\phi\textrm{Update}(x_i,\hat{y}_i, y_i^*, \theta_{t}; \phi)$. only once for each $t$, rather than recomputing these gradients for all subsequent time steps. 
These choices allow the memory use of our training algorithm to remain constant in $r$.
We make the same choice for our $K$ looped steps in a single application of the Update function, so the gradient for the update at step $k$ depends only on $g_\phi(x_i,\hat{y}_i, y_i^*, \theta^{(k)})$ and not $\theta^{(k-1)}$. See Fig. \ref{fig:memory_combined} for a graph of memory use depending on $r$ and $k$. 

\vspace{2pt}
\noindent\textbf{Experiment runtimes.} We now give runtimes for experiments in the paper.
Building the belief graphs takes 25 hours for FEVER ($n=10,444$) and 17.5 hours for LeapOfThought ($n=8642$) on an NVIDIA RTX 2080 GPU. Computing summary statistics for graphs takes 3 hours on FEVER and 3 hours for LeapOfThought for statistics besides Update-Transitivity. We compute Update-Transitivity for LeapOfThought with a subset of 4000 points, which takes 45 hours. 

All other experiments are run on a NVIDIA V100 32GB GPU. Training the task models takes 7 minutes for LeapOfThought, 45 minutes for FEVER, 4 hours for zsRE, and 10 hours for Wikidata5m. Training the learned optimizer with $r=1$ takes 2.3 hours for LeapOfThought, 5 hours for FEVER, 9.5 hours for zsRE, and 16 hours for Wikidata5m. Training the learned optimizer with $r=10$ takes 53 minutes for LeapOfThought, 2.9 hours for FEVER, 7 hours for zsRE, and 12.5 hours for Wikidata5m. Computing update statistics with the off-the-shelf optimizers with $r=1$ takes 4 minutes for LeapOfThought, 30 minutes for FEVER, 2.3 hours for zsRE, and 3.9 hours for Wikidata5m. With $r=10$, the baselines require 1 minute for LeapOfThought, 15 minutes for FEVER, 54 minutes for zsRE, and 1.8 hours for Wikidata5m. Total runtimes for each experiment should take into account multiple conditions and multiple seeds of each model being run.

\subsection{Hyperparameters and Objective Terms.}
\label{app:hyperparameters}

\vspace{2pt}
\noindent\textbf{Training hyperparameters}. We fit our RoBERTa-base and BART-base task models to their respective datasets with the following hyperparameters: We train for 10 epochs on the binary tasks, and 20 for the sequence-to-sequence tasks. When predicting with BART-base, we use a beam search with width 5. In each case, we use AdamW from \texttt{torch.optim} with a LR of 1e-5 and weight decay of 1e-4. We select the best model according to the best dev set accuracy, checkpointing after each training epoch. The learned optimizers are optimized with AdamW, using a learning rate of 3e-4 and weight decay of 0. We train the learned optimizer for 5 epochs on each dataset except for LeapOfThought, which we train for 10 epochs given its smaller size. The learned optimizers are also selected based on dev set performance, with checkpointing after each training epoch. Their selection criterion is a raw average of Update Success Rate (averaged over each kind of data), Retain Rate (\emph{Local Neutral}) and $\Delta$-Acc, with terms dropped when they cannot be computed given the available data. Note that dev epochs with zsRE and Wikidata5m are fairly slow, so in order to speed up our experiments we compute dev epochs with a subset of 4000 dev points. 

\vspace{2pt}
\noindent\textbf{Learned optimizer}. We give the final hyperparameter and objective terms used in each experiment in Table \ref{tab:method-hyperparameters}. Our objective ablation is given in \ref{tab:objective_term_ablation}, and we select the best performing condition for each dataset according to dev set performance, using the same selection criterion outlined previously. We keep all weight coefficients $\lambda_i$ equal rather than tuning them. Main refers to the first term in Eq. \ref{eq:update_objective}, plus the KL term with random data. We use $K_\textrm{train}\leq5$ for all experiments. For results across $K$ values on zsRE, see Fig. \ref{fig:k_ablation}.

\begin{table}
    \centering
    \small
    \begin{tabular}{l c c c}
    \toprule
    Dataset & $r_\textrm{test}$ & $k$ & Objective \\
    \midrule
      \multirow{2}{*}{FEVER}
           & 1 & 5 & Main \\
           & 10 & 1 & Main \\
       \midrule
      \multirow{2}{*}{LeapOfThought}
      & 1 & 5 & Main \\
       & 10 & 1 & Main \\
       \midrule
      \multirow{2}{*}{zsRE}
      & 1 & 5 & Main \\
                & 10 & 5 & Main \\
        \midrule
      \multirow{2}{*}{Wikidata5m}    & 1 & 5 & Main+Para \\
          & 10 & 5 & Main+Para \\
     \bottomrule
    \end{tabular}
    \vspace{-2pt}
    \caption{Final hyperparameters and objective terms of the learned optimizer for each task.}
    \vspace{-5pt}
    \label{tab:method-hyperparameters}
\end{table}

\begin{table}[t]
\begin{center}
\small
\begin{tabular}{l S}
\toprule 
Relation & {\% Test Data} \\
\midrule
Place of Birth & 11.00\\
Award Received & 11.00\\
Cause of Death & 5.66\\
Place of Death & 11.00\\
Place of Burial & 8.33\\
Educated At & 11.00\\
Child & 11.00\\
Occupation & 11.00\\
Spouse & 11.00\\
Sibling & 9.01\\
\bottomrule
\end{tabular}
\end{center}
\vspace{-10pt}
\caption{Wikidata relations and their proportion of the test data.}
\vspace{-10pt}
\label{tab:wikidata-relations}
\end{table}

\vspace{2pt}
\noindent\textbf{Baseline update method}. We tune a baseline off-the-shelf optimizer separately for each dataset, using $r_\textrm{test}=1$. Our performance criterion is the same as with learned optimizers, a raw average of Update Success Rate (averaged over each kind of data), Retain Rate (\emph{Local Neutral}) and $\Delta$-Acc. The grid search is over the following parameters: The off-the-shelf optimizers are from \texttt{torch.optim} and include \{AdamW, SGD, and RMSProp\} with default arguments (except for the learning rate). We consider a number of maximum steps in \{5, 10, 100\}. The learning rates we consider depend on the optimizer: \{1e-4, 1e-5, 1e-6\} for AdamW, \{1e-4, 1e-5, 1e-6\} for RMSProp, and \{1e-1, 1e-2, 1e-3\} for SGD. The LR ranges were selected after some initial manual exploration of the space. Our final hyperparameter values are shown in Table \ref{tab:baseline-hyperparameters} for each dataset.
For comparison, \citet{de2021editing} use RMSProp with 100 update steps. The LR for zsRE and Wikidata5m may seem quite high, but this is the condition that actually does the least damage to the model's accuracy on other data, $\Delta$-Acc. The baseline optimizes all of the trainable parameters in the language model, unlike the learned optimizer which optimizes only attention and feedforward weights for purposes of parameter efficiency. 

\begin{table}
    \centering
    \small
    \begin{tabular}{l c c c}
    \toprule
    Dataset & Optimizer & LR & Num. Steps \\
    \midrule
      FEVER         & AdamW & 1e-6 & 100 \\
      LeapOfThought & SGD & 1e-2 & 100 \\
      zsRE          & SGD & 1e-1 & 10 \\
      Wikidata5m    & SGD & 1e-1 & 10 \\
     \bottomrule
    \end{tabular}
    \vspace{-2pt}
    \caption{Final hyperparameters of the baseline update method for each task.}
    \vspace{-8pt}
    \label{tab:baseline-hyperparameters}
\end{table}

\begin{table*}
    \centering
    \small
    \begin{tabular}{p{.36\textwidth} p{.3\textwidth} p{.24\textwidth}}
    \toprule
    Ours & \citet{de2021editing} & \citet{mitchell2021fast} \\
    \midrule
         Update Success Rate (\emph{Main Input}) & Success rate & Edit success  \\
         Update Success Rate (\emph{Paraphrase}) & Equivalence accuracy & Edit success \\
         Update Success Rate (\emph{Entailed Data}) & - & - \\
         Retain Rate (\emph{Local Neutral}) & - & - \\
         Retain Rate (\emph{All Data}) & Retain accuracy  & -  \\
         $\Delta$-Acc (\emph{All Data}) & Performance deterioration & Drawdown \\
     \bottomrule
    \end{tabular}
    \vspace{-6pt}
    \caption{A glossary of terms used in work on model update methods. Note metrics are not always calculated in exactly the same way. For instance, Performance deterioration is a ratio in accuracies rather than difference in accuracies, and edit success from \citet{mitchell2021fast} combines two metrics in our case. The performance metric in \citet{zhu2020modifying} is an average of Update Success Rate (\emph{Main Input}) and $\Delta$-Acc.}
    \vspace{-5pt}
    \label{tab:metrics-glossary}
\end{table*}

\subsection{Wikidata5m Additional Details.}

We construct four paraphrases per Main Input by selecting from a set of alternative phrasings for the entity and relation in the Main Input. The syntax for each paraphrase follows the same simple template as the Main Input, in contrast to zsRE where syntax differs between paraphrases. A couple details remain. Some relations are one-to-many, and therefore we accumulate valid completing entities from the data as possible answers; later we compute accuracy as an exact match with any possible answer. All 10 relations appear in each split of the data. Only 33.80\% and 37.18\% of the entities in the dev and test splits are seen in the training data, though we do not find that models perform better on entities seen in training.

\subsection{LeapOfThought Additional Details}

The LeapOfThought dataset consists of a fact and a claim for each datapoint, where the truth of the fact implies that the claim has label $y_i$ (True/False). All of the facts in the data are true, while half of the claims are true and half are false. When training the learned optimizer, we treat the the facts as the Main Input when training the learned optimizer and claims as entailed data. When training the True/False classifier, we fit to the claims, for which test accuracy is 83.65 ($\pm$ 1.05). This seems to generalize well to the facts, as test accuracy here is 93.66 ($\pm$0.87), although as the low contrapositive accuracy suggests (Table \ref{tab:belief_metrics_by_correctness}), the model seems to be too prone to predicting true for this data. 

Since very few of the Main Inputs are predicted as false, we run into a small dilemma when fitting the learned optimizer with the use of the entailed data objective term. The entailment between fact and claim only holds when the fact is true, so we can only compute the objective when updating a point from false to true. This ends up being less than 10\% of the training data. We ultimately choose to oversample points that fit this description during training of the learned optimizer, which allows the learned optimizer to fully fit to the entailed data. 
Also note that during learned optimizer training, we include Entailed Data \emph{from other data points besides the Main Input} in the KL term in Eq. \ref{eq:update_objective}, and we measure $\Delta$-Acc using both Main Inputs and Entailed Data. 

\begin{table*}[t]
\small
\begin{center}
\begin{tabular}{l l c c c c}
\toprule
{Dataset}&{Model}&{Acc}&{Paraphrase Cons $\uparrow$}&{Entailment Acc $\uparrow$}&{Contrapositive Acc $\uparrow$} \\
    \midrule
    FEVER   &   RoBERTa-base   &   78.29 (0.86)    &   -   &   -  &   -  \\
    LeapOfThought   &   RoBERTa-base   &   93.66 (0.87)   &   -  & 85.63 (1.08) &   16.51 (2.71) \\
    zsRE   &   BART-base   &   21.01 (0.64)    &   69.50 (1.09)   &   - &   -   \\
    Wikidata5m   &   BART-base   &   10.21 (0.59)    &   25.84 (0.53)   &   - &   -     \\
    \bottomrule
\end{tabular}
\end{center}
\vspace{-7pt}
\caption{Model accuracy and belief metric results and for four datasets.}
\vspace{-2pt}
\label{tab:belief_metrics_appendix}
\end{table*}

\begin{table*}[t]
\begin{center}
\small
\begin{tabular}{l l  p{0.36\textwidth} p{0.26\textwidth}}
\toprule 
Dataset & Data Type & Input & Label(s)  \\
\midrule
\multirow{5}{*}{zsRE}
& Main Input & What did Gifford Pinchot die of? & \multirow{2}{0.26\textwidth}{\{Leukemia\}} \\
& Paraphrase & How did Gifford Pinchot die? & \\
\addlinespace[6pt]
& Main Input & Player Ali Kanaan plays for what team? & \multirow{2}{0.26\textwidth}{\{Sporting Al Riyadi Beirut\}} \\
& Paraphrase & What team is Ali Kanaan associated with? &  \\
\midrule
\multirow{10}{*}{Wikidata5m}
& Main Input & Margarita Nolasco Armas has relation `place of birth' to & \multirow{3}{0.26\textwidth}{\{Orizaba, Veracruz; Orizaba; etc.\}} \\
& Paraphrase & SusunW/Margarita Nolasco Armas has relation `born at' to & \\
\addlinespace[1pt]
& Local Neutral & Margarita Nolasco Armas has relation `place of death' to & Mexico City; Ciudad de Mexico; etc. \\
\addlinespace[2pt]
& Main Input & Mary Good has relation `award received' to & \multirow{2}{0.26\textwidth}{\{Garvan-Olin Medal; Arkansas Women's Hall of Fame; etc.\}} \\
& Paraphrase & Mary Lowe Good has relation `winner of' to &  \\
\addlinespace[1pt]
& Local Neutral & Mary Good has relation `educated at' to & \{The University of Arkansas; U Arkansas; etc.\} \\
\midrule
\multirow{2}{*}{FEVER}
& Main Input & Tardigrades are also known as space bears. & True \\
& Main Input & The Lion belongs to the genus Vulpes. & False \\
\midrule
\multirow{5}{*}{LeapOfThought}
& Main Input & A viper is a vertebrate. & True \\
& Entailed Data & A viper has a brain. & True \\
\addlinespace[6pt]
& Main Input & A amaranth is a herb. & True \\
& Entailed Data & A amaranth has a nose. & False \\
\toprule
\end{tabular}
\end{center}
\vspace{-10pt}
\caption{Example datapoint from each dataset, and auxiliary data that accompanies the Main Input.}
\vspace{-2pt}
\label{tab:data-examples-appendix}
\end{table*}

\section{Noise in Datasets}
\label{app:noise-in-datasets}

We briefly document some shortcomings of each dataset, with reference to examples in Table \ref{tab:data-examples-appendix}.

\vspace{2pt}
\noindent\textbf{FEVER.} Some claims are slightly vague or ambiguous when taken on their own. For instance ``Doug Ducey was the CEO of Cold Stone Creamery and offered many opportunities to new hires" is rated True, though this will depend heavily on what one thinks ``many opportunities" means. Similar whether or not ``L.A. Guns is a tattoo shop" depends on which ``L.A. Guns" one is referring to, the tattoo shop or metal band. Of course, this is a generic issue of language, and not unique to this dataset. Some inputs seem to be a matter of person opinion: ``Los Angeles is known for its food" is rated False. 

\vspace{2pt}
\noindent\textbf{LeapOfThought.} Many examples use an ``is a" relation, producing sentences like ``A sunlight is a good health." This could be more false than true, but it's a fairly nonsensical statement to begin with. There are also other nonsensical or vague examples in the data: ''A friar is the opposite of mineral" is labeled False. ``A detective desires equal opportunity." is labeled True. It is not immediately clear what conditions would make these statements true or false.

\vspace{2pt}
\noindent\textbf{zsRE.} Some questions invoke potentially one-to-many or temporally dependent relations, though there is only one ground-truth answer per question in this dataset. For instance, a paraphrase of the question about Gifford Pinchot in Table \ref{tab:data-examples-appendix} is: "What disease did Gifford Pinchot have?" A person might have had many diseases over their life which could all be valid responses. The answer is especially ambiguous for spatial relations, where a valid answer might refer to a city, region, country, province, or continent. 

\vspace{2pt}
\noindent\textbf{Wikidata.} Aliases sometimes vary greatly even as they refer to the same person, or they are simply noisy. For example, as shown in Table \ref{tab:data-examples-appendix}, ``SusunW" appears in an entity name, but this is actually a username of someone who contributed to the Wikipedia article for Margarita Nolasco Armas. Meanwhile, other aliases for J.R.R Tolkien include ``Tolkienian" and ``Mabel Suffield," his mother. Rephrasings of relations might also create confusing inputs, e.g. switching ``child" with ``has kids," ``daughter", or ``son." Similar to zsRE, some relations are also one-to-many and temporally dependent (like occupation), though we hope that by using many valid answers we circumvent this issue to some extent when calculating prediction correctness. 

\begin{figure*}[t]
    \centering
    \includegraphics[width=0.89\textwidth]{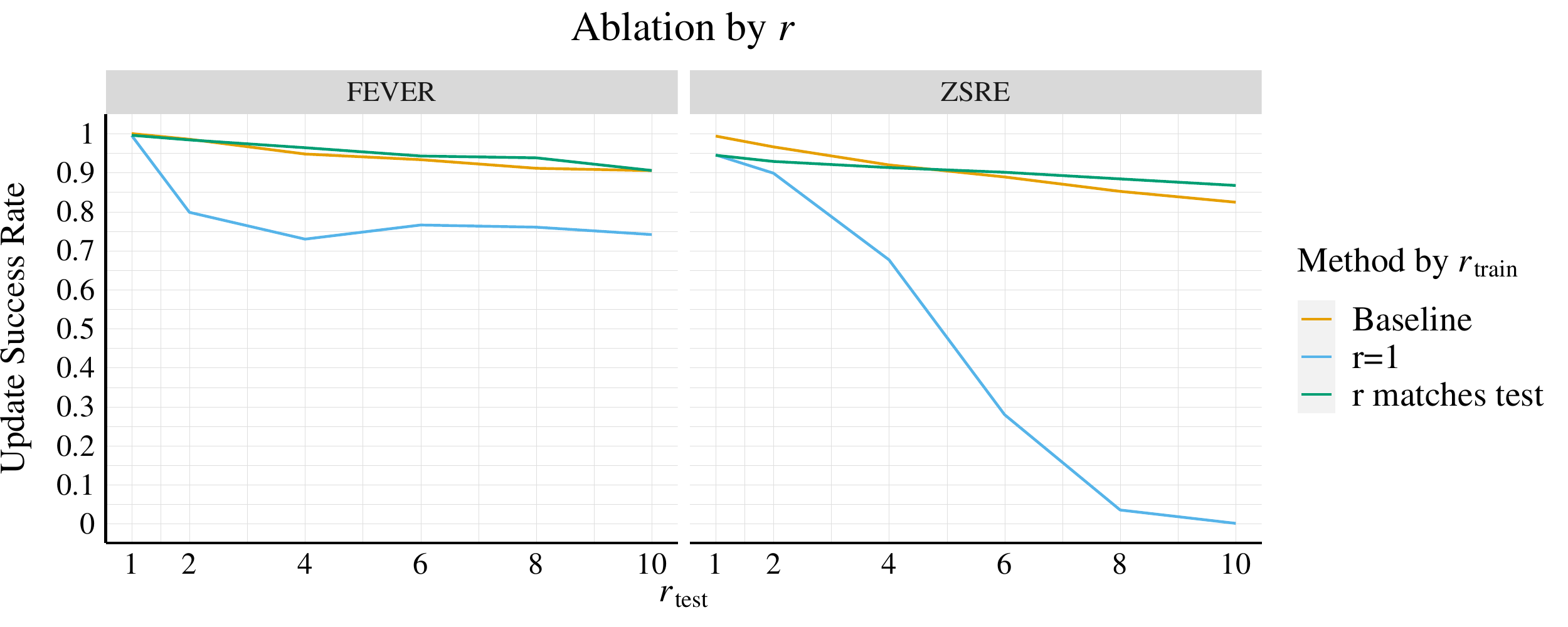}
    \vspace{-9pt}
    \caption{Ablation across values of $r$ for training and testing. On zsRE, our method outperforms the baseline when $r_\textrm{test}=10$, and the gap is likely to increase as $r_\textrm{test}$ rises further. When using a non-sequential objective from past work, performance declines drastically as $r_{\textrm{test}}$ rises.
    }
    \label{fig:r_ablation}
    \vspace{-6pt}
\end{figure*}

\section{Metric Computation and Bootstrap Details}
\label{app:metric_and_bootstrap_details}

\noindent\textbf{Metric computation.} The only computationally difficult metric to calculate is $\Delta$-Acc, which requires computing the updated language model's accuracy on other data after every single belief update. We randomly sample other data after every update for this purpose, using $n=30$ points for zsRE and Wikidata5m and $n=200$ points for FEVER and LeapOfThought. We ensure that all evaluation data is used at some point during this sampling by preferentially selecting data that has been infrequently selected before. We note that paraphrase consistency is easy to evaluate for a small number of paraphrases per datapoint, as we have for both zsRE and Wikidata5m. Additionally, on LeapOfThought, we compute $\Delta$-Acc using both Main Inputs and Entailed Data.

\noindent\textbf{Update-Transitivity caveat.} The \% Update-Transitivity metric represents the answer to the question: if updating belief A changes belief B, and updating belief B changes belief C, what proportion of the time does updating A change C? We would treat this as a normative metric that we hope to maximize, except we do not know in general whether there is a confounding belief D that determines the relationship between B and C. If changing A also changed a confounding belief D, then we might not be able to expect that C should change too. 
That said, when we have no reason to think there are such confounding beliefs, we would expect a logically consistent model to display 100\% Update-Transitivity of their beliefs. In Fig. \ref{fig:belief-graph}, for instance, we see no reason to suspect there are confounding beliefs for the relationship between the date Bessie Smith died and the writer of Despicable Me 2, and therefore we would expect that updating the belief about what album Hot Right Now is on would change the belief in Despicable Me 2's authorship (which it does). 

\noindent\textbf{Bootstrap computation.} We account for sample and seed variance by block bootstrap \cite{efron1994introduction}. When there is a single statistic per data point, like Main Input Update Success, we form a matrix of shape $n \times s$ for $n$ data points and $s$ model seeds (where the seed was used for both task model training and learned optimizer training). We then resample rows and columns of this matrix 10,000 times, which was sufficient for convergence. When we perform hypothesis tests for the difference in statistics between conditions, we pair the data points by using the same rows of this matrix at each step of the bootstrap (i.e. we conduct paired tests). For metrics involving multiple data points per Main Input, like paraphrases or other random data, we make a simplifying assumption where we do not resample the multiple data points but just compute the average metric for those data points and treat that as the ground-truth statistics for the Main Input. We explored using a full 3-dimensional bootstrap, where we resample among these extra datapoints by constructing a matrix of shape $n \times s \times n$, but it was quite slow and gave similar results to the block bootstrap. 

\section{Additional Results}
\label{app:additional-results}

\begin{table}[t]
\small
\begin{center}
\begin{tabular}{l c c c}
\toprule
& \multicolumn{2}{c}{Update Success Rate}  & $\Delta$-Acc \\
\cmidrule(lr){2-3} \cmidrule(lr){4-4}
Desired Label & Main Input & Paraphrases & All Data  \\
\midrule
Beam Label & 91.19 (0.5) & 92.07 (0.8) & -0.39 (0.1) \\
Hard Label & 94.46 (0.7) & 94.45 (0.7) & -0.24 (0.1) \\
\bottomrule
\end{tabular}
\end{center}
\vspace{-5pt}
\caption{
Update metrics by optimizer training labels.
}
\vspace{-5pt}
\label{tab:label_ablation}
\end{table}

\begin{table*}[t]
\small
\begin{center}
\begin{tabular}{l l c c c c c r}
\toprule
\multicolumn{2}{l}{\textbf{Objective Term Ablation}} & \multicolumn{3}{c}{Update Success Rate}  & \multicolumn{2}{c}{Retain Predictions} & \multicolumn{1}{c}{$\Delta$ Acc} \\
\cmidrule(lr){3-5} \cmidrule(lr){6-7} \cmidrule(lr){8-8}
Dataset & Objective & Main Input & Paraphrases & Entailed Data & Local Neutral         & All Data        & \multicolumn{1}{c}{All Data}  \\
\midrule
\multirow{2}{*}{FEVER} 
 & Main & 100 (0.0)
       & -           & -          & -                    & 98.27 (0.1)            & -0.15 (0.1)       \\
& \hspace{1pt} (no KL) & 100 (0.0)
       & -           & -          & -                    & 40.42 (0.6)             & -27.19 (1.2)      \\
\midrule
\multirow{2}{*}{LeapOfThought} & Main & 100 (0.0)       & -           & 76.43 (5.3)          & -                    & 96.84 (0.3)            & -1.22 (0.8)      \\
 & \hspace{0pt}+Ent & 100 (0.0)       & -           & 71.87 (5.3)          & -                    & 96.52 (0.3)            & -0.40 (0.8)      \\

\midrule
\multirow{2}{*}{zsRE}            & Main     & 94.46 (0.4)       & 94.44 (0.7)        & -             & -                    & 81.96 (0.4)            & -0.24 (0.1)      \\
            & \hspace{0pt}+Para     & 93.75 (0.4)       & 94.41 (0.7)        & -             & -                    & 75.24   (0.5)         & -0.42 (0.2)      \\
\midrule

\multirow{4}{*}{Wikidata5m} 
& Main     & 88.67 (0.7)       & 64.12 (0.7)        & -             & 49.78 (1.0)                 & 71.04 (0.5)            & -1.54 (0.3)   \\  
& \hspace{0pt}+Para     & 87.46 (0.7)        & 81.06 (0.7)        & -             & 47.15 (1.0)                 & 63.02 (0.6)            & -1.55 (0.3)   \\  
& \hspace{0pt}+LN     & 87.73 (0.7)       & 59.75 (0.7)       & -             & 60.49 (1.0)                 & 72.69 (0.6)           & -1.57 (0.3)  \\
& \hspace{0pt}+Para+LN     & 87.02 (0.7)       & 81.18 (0.7)        & -             & 56.86 (1.0)                 & 68.42 (0.6)            & -1.65 (0.3)   \\  

\bottomrule
\end{tabular}
\end{center}
\vspace{-8pt}
\caption{Belief update results by the objective terms used for the learned optimizer. We do not bold any numbers based on statistical significance. For tuning purposes we select whichever condition achieves the higher selection criterion without testing for statistical significance.}
\vspace{-4pt}
\label{tab:objective_term_ablation}
\end{table*}

\vspace{2pt}
\noindent\textbf{Ablation across num. sequential steps.} Fig. \ref{fig:r_ablation} shows the results for an ablation across $r_\textrm{test}$ using two kinds of learned optimizers: SLAG$_1$, where $r_{\textrm{train}}=1$, and a SLAG condition where $r_{\textrm{train}}=r_{\textrm{test}}$. It is critical to the success of learned optimizers to train them to update points sequentially when this is a desired application. Further, sequential updating with sequence prediction tasks is the only setting where we see learned optimizers outperform baselines across all relevant metrics. 

\vspace{2pt}
\noindent\textbf{Choosing training labels for learned optimizers.} In early experiments, we found that it is beneficial to use all data points (including correctly predicted points) as Main Inputs \emph{during training}, rather than restricting training to only incorrectly predicted points. 
We still focus on correcting wrong outputs at test time. But so we must select what label to use during optimizer training. 
To get a Hard Label, we use the correct label for incorrectly predicted points, and for correctly predicted points, we simply draw a label randomly from the labels in the training data. The alternative Beam Label condition uses a sample from the model's beam search for a data point, as done in past work \cite{de2021editing, mitchell2021fast}. We show update metrics for zsRE split by the desired label in Table \ref{tab:label_ablation}.
If one's goal is to fix wrong model outputs, then it is much better to use either the correct label or a random label as the desired model output during training rather than a sample from the model's beam search. Update success improves by 3.27 ($\pm$0.65; $p{<}$\num{1e-4}) points for the Main Input and 2.38 ($\pm$1.05; $p{<}$\num{1e-4}) for Paraphrases, while $\Delta$-Acc rises by 0.15 ($\pm$0.18; $p{=}.09$).

\begin{figure}[t]
    \centering
    \includegraphics[width=0.45\textwidth]{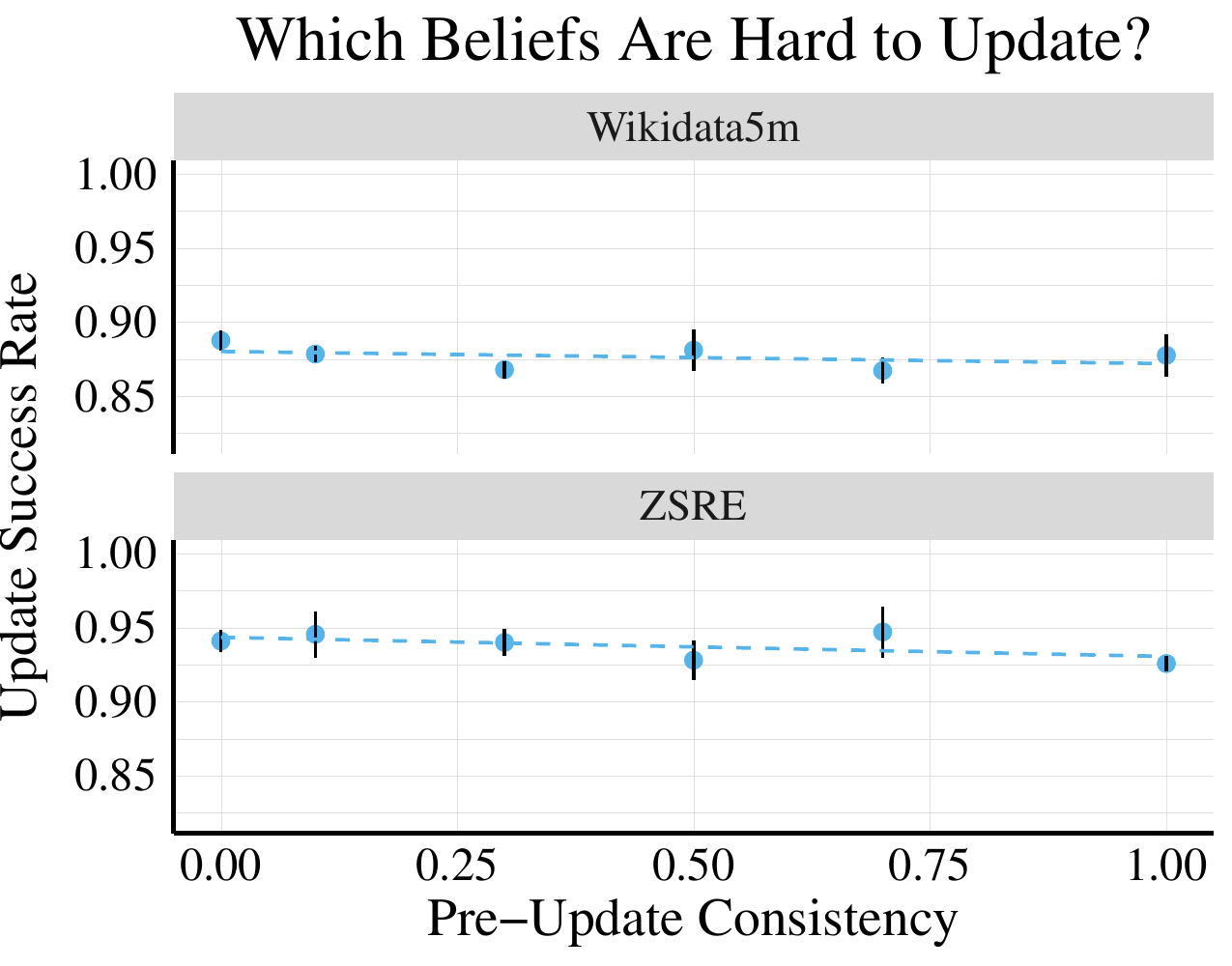}
    \vspace{-8pt}
    \caption{Beliefs are neither easier nor harder to update depending on their consistency beforehand.}
    \label{fig:upd-suc-by-before-cons}
    \vspace{-2pt}
\end{figure}

\vspace{2pt}
\noindent\textbf{Which beliefs are hard to update?} We hypothesize that beliefs will be easier to update when they are more belief-like to begin with.
We principally measure this via the correlation between update success rate and a belief's consistency on paraphrases before the update, for our learned optimizer in a single-update setting ($r=1$). 
Surprisingly, we observe no relationship between update success and the belief consistency. The correlation between consistency and update success is near 0 for both zsRE ($\rho=-.027$) and Wikidata5m ($\rho=.013$); see Fig. \ref{fig:upd-suc-by-before-cons} for a plot of the relationship.
So it appears that the learned optimizer can update model beliefs independently of how belief-like they are to begin with. 
We would also be interested in considering consistency under entailment, but the update success rate on LeapOfThought is already 100\%, so there is no variance to explain. 

\begin{figure}[t]
    \centering
    \includegraphics[width=0.45\textwidth]{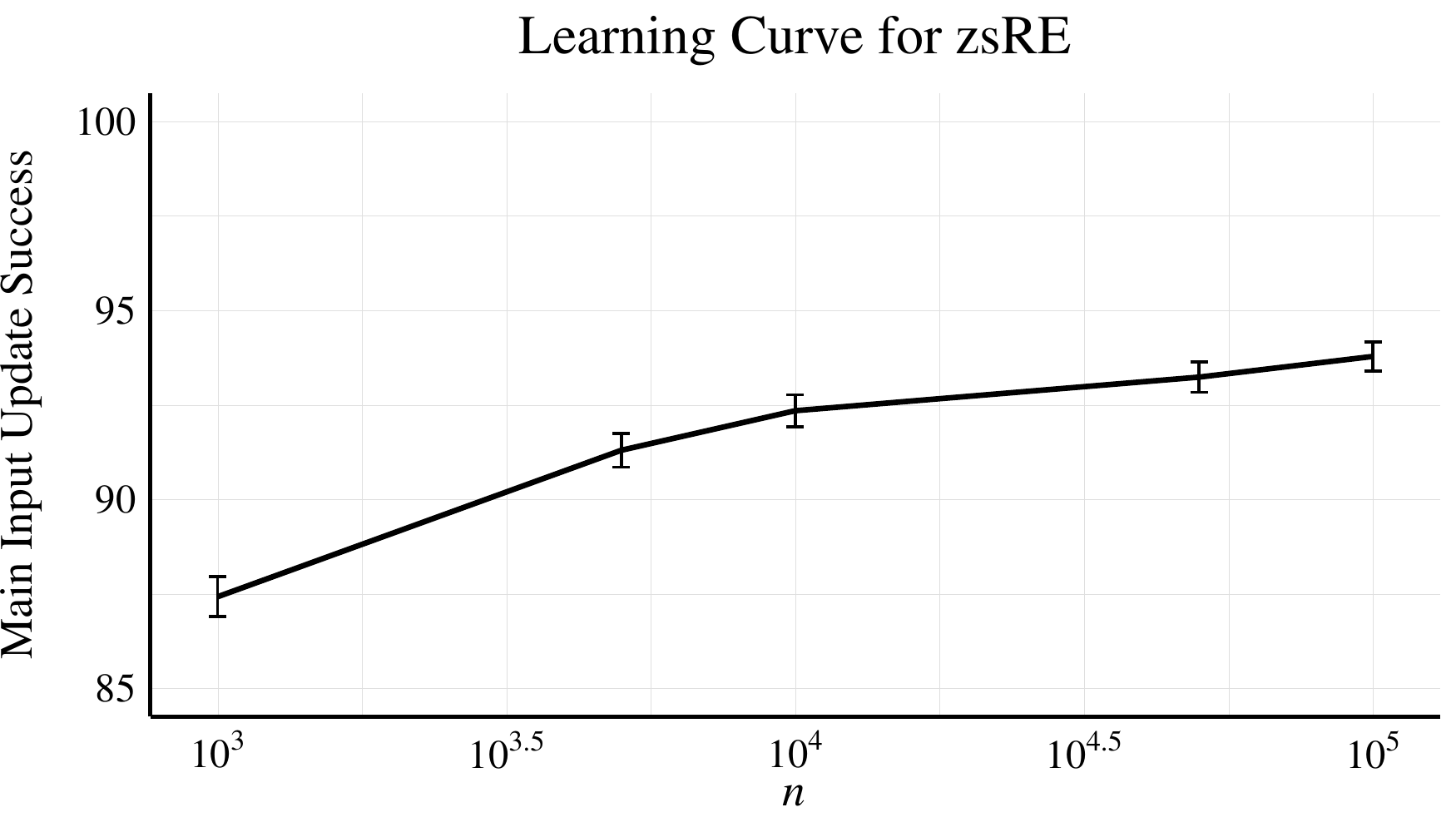}
    \vspace{-12pt}
    \caption{Main Input Update Success Rate across training set sizes, using SLAG on zsRE.}
    \label{fig:learning_curve}
    \vspace{-8pt}
\end{figure}

\vspace{2pt}
\noindent\textbf{Learning curve.} In Fig. \ref{fig:learning_curve} we show the learning curve of a learned optimizer trained with SLAG on zsRE. The Main Input Update Success Rate steadily rises as a function of the training set size.

\begin{figure}
    \centering
    \includegraphics[width=0.49\textwidth]{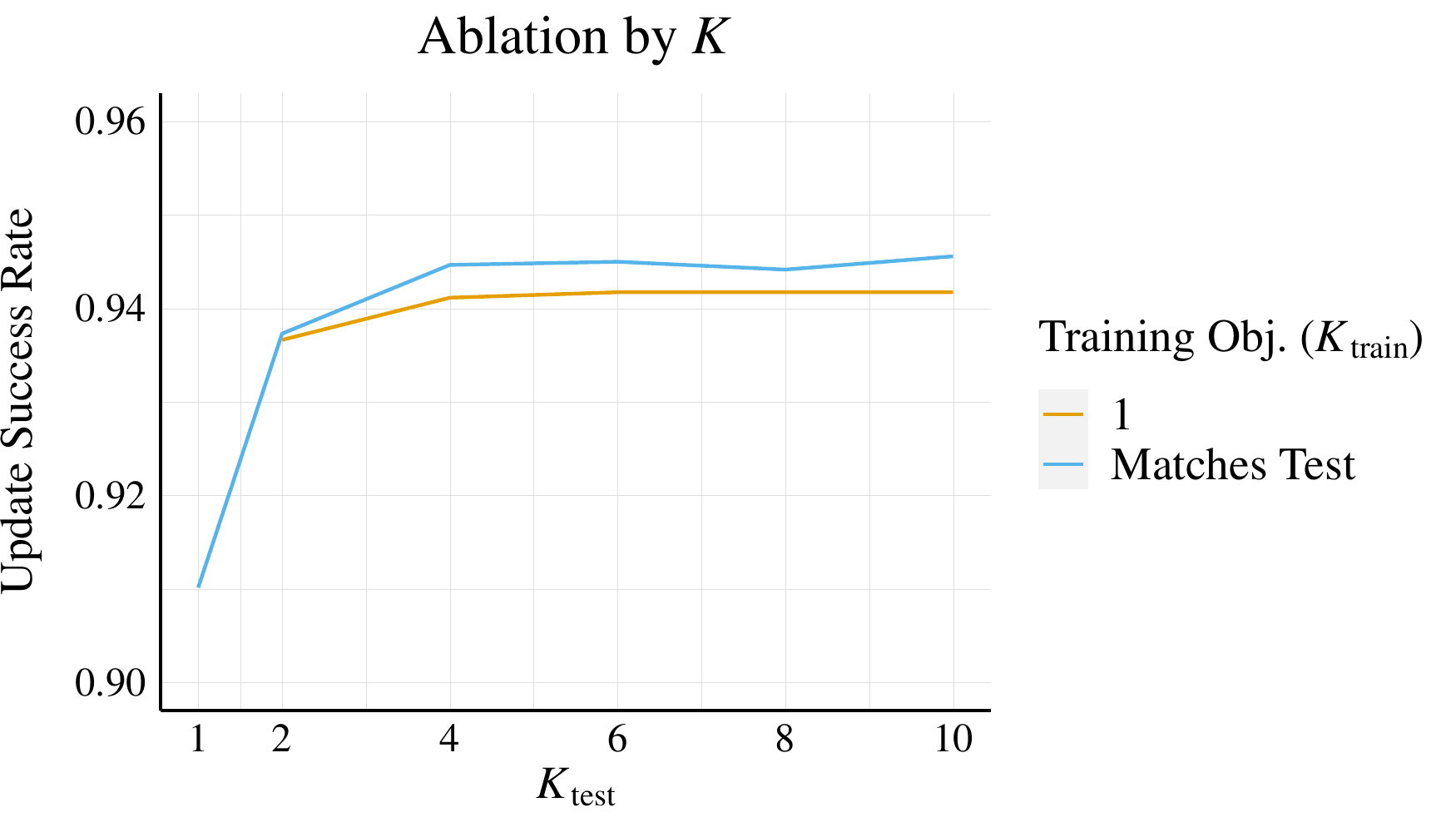}
    \vspace{-9pt}
    \caption{Ablation across values of $K$ for training and testing, using SLAG on zsRE. It is useful to train the optimizer using the value of $K$ it will use at test time.}
    \label{fig:k_ablation}
    \vspace{-4pt}
\end{figure}

\vspace{2pt}
\noindent\textbf{Ablation by num. update steps.} Fig. \ref{fig:k_ablation} shows the results of an ablation across values of $K$ using a learned optimizer trained using SLAG with $r=1$ on zsRE. Main Input Update Success rises by over three points by increasing $K_\textrm{test}$ from 1 to at least 5. Using a value of $K_\textrm{train}$ that matches $K_\textrm{test}$ gives a further increase of about 0.5 points.

\begin{figure}
    \centering
    \includegraphics[width=0.49\textwidth]{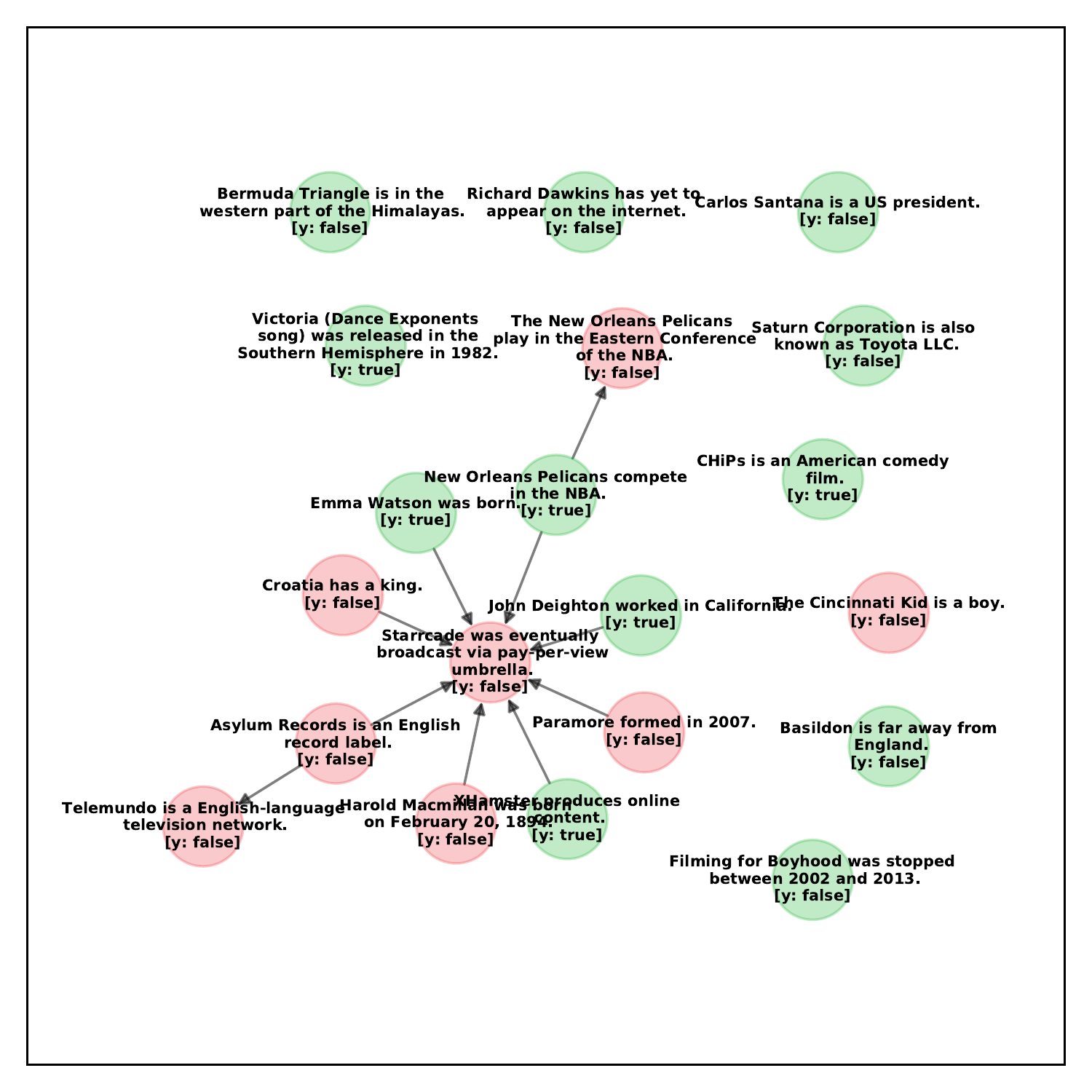}
    \vspace{-12pt}
    \caption{A random subgraph of the belief graph for FEVER. Note all nodes actually are connected to at least one another node.}
    \label{fig:random-subgraph-0}
    \vspace{-8pt}
\end{figure}

\begin{figure}
    \centering
    \includegraphics[width=0.49\textwidth]{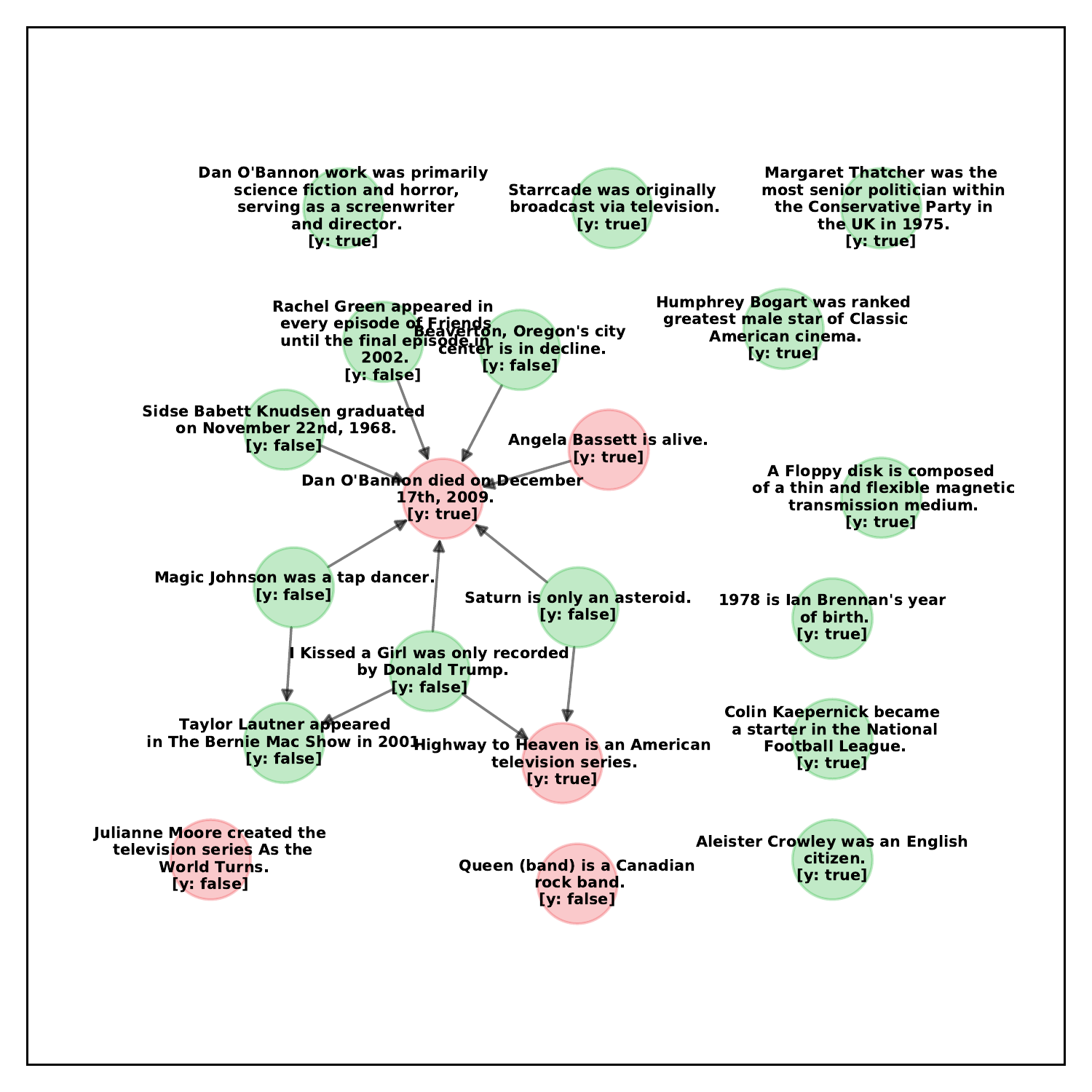}
    \vspace{-12pt}
    \caption{A random subgraph of the belief graph for FEVER. Note all nodes actually are connected to at least one another node.}
    \label{fig:random-subgraph-1}
    \vspace{-8pt}
\end{figure}

\begin{figure}
    \centering
    \includegraphics[width=0.49\textwidth]{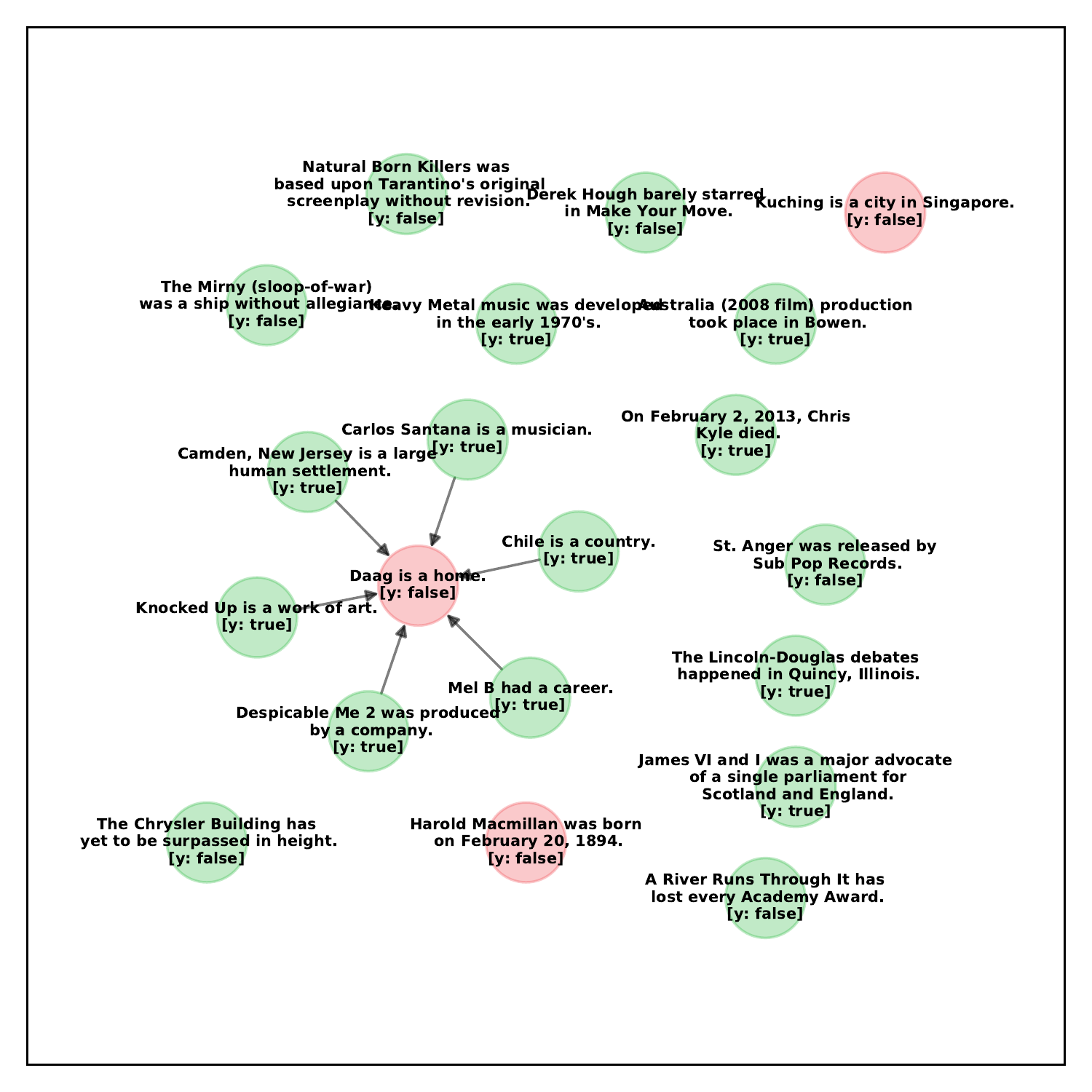}
    \vspace{-12pt}
    \caption{A random subgraph of the belief graph for FEVER. Note all nodes actually are connected to at least one another node.}
    \label{fig:random-subgraph-2}
    \vspace{-8pt}
\end{figure}

\end{document}